\documentclass[11pt, 3p]{elsarticle}

\usepackage[utf8]{inputenc}
\usepackage{lmodern}
\usepackage{booktabs}
\usepackage{amsfonts}
\usepackage{nicefrac}
\usepackage{microtype}
\usepackage{subfig}
\usepackage{caption}
\usepackage{graphicx}
\usepackage{tikz}  
\usetikzlibrary{shapes,arrows,positioning,fit,backgrounds}  
\usepackage{standalone}  
\usepackage{import}
\usepackage[numbers]{natbib}
\usepackage{url}
\usepackage{amsmath}
\usepackage{algorithm}
\usepackage{algorithmic}
\usepackage{amssymb}
\usepackage{float} 
\usepackage{booktabs}
\usepackage{siunitx}
\usepackage{hyperref}

\makeatletter
\def\@journal{}
\def\ps@pprintTitle{%
  \let\@oddhead\@empty
  \let\@evenhead\@empty
  \let\@oddfoot\@empty
  \let\@evenfoot\@oddfoot
}
\makeatother

\begin{document}
\setlength{\textfloatsep}{10pt plus 2pt minus 2pt}

\begin{frontmatter}

\title{A Graph-Enhanced Deep-Reinforcement Learning Framework for the Aircraft Landing Problem}

\author[add]{Vatsal K.~Maru\corref{cor}}
\ead{vatsal.maru@utdallas.edu}

\address[add]{Information Systems, Jindal School of Management, The University of Texas at Dallas, 800 W Campbell Rd, JSOM Office 4.414, Richardson, TX 75080, USA.}

\cortext[cor]{Corresponding author. ORCID: 0000-0002-8210-879X}

\begin{abstract}
The Aircraft Landing Problem (ALP) is one of the challenging problems in aircraft transportation and management. The challenge is to schedule the arriving aircraft in a sequence so that the cost and delays are optimized. There are various solution approaches to solving this problem, most of which are based on operations research algorithms and meta-heuristics. Although traditional methods perform better on one or the other factors, there remains a problem of solving real-time rescheduling and computational scalability altogether. This paper presents a novel deep reinforcement learning (DRL) framework that combines graph neural networks with actor-critic architectures to address the ALP. This paper introduces three key contributions: A graph-based state representation that efficiently captures temporal and spatial relationships between aircraft, a specialized actor-critic architecture designed to handle multiple competing objectives in landing scheduling, and a runway balance strategy that ensures efficient resource utilization while maintaining safety constraints. The results show that the trained algorithm can be tested on different problem sets and the results are competitive to operation research algorithms. The experimental results on standard benchmark data sets demonstrate a 99.95\% reduction in computational time compared to Mixed Integer Programming (MIP) and 38\% higher runway throughput over First Come First Serve (FCFS) approaches. Therefore, the proposed solution is competitive to traditional approaches and achieves substantial advancements. Notably, it does not require retraining, making it particularly suitable for industrial deployment. The framework’s capability to generate solutions within $<$1 second enables real-time rescheduling, addressing critical requirements of air traffic management.
\end{abstract}

\begin{keyword}
aircraft landing problem (ALP) \sep deep reinforcement learning (DRL) \sep graph neural networks (GNN) \sep air traffic management (ATM) \sep runway optimization \sep real-time scheduling
\end{keyword}

\end{frontmatter}

\section{Introduction}\label{sec:introduction}
The aircraft landing problem (ALP) is one of the scheduling problems in the sequencing of aircraft on runways. The ALP focuses primarily on arriving aircraft. With global air traffic projected to exceed 9 billion passengers by 2025 \cite{ICAO2025forecast}, and major hubs like Hartsfield-Jackson Atlanta International Airport handling nearly 96 million passengers annually \cite{Zhou2018}, efficient runway utilization has become essential to maintain operational safety and minimize delays. The resulting air traffic on runways \cite{AirportsCouncilInternational} leaves airports with two possible options: expanding the infrastructure or optimizing the existing infrastructure to meet the growing demand. Even if airports expand the infrastructure, it is not a quick process to build runways, which leaves airport authorities and Air Traffic Control (ATC) with optimization problems. A frequent scheduling technique the ATC uses is First Come First Serve (FCFS) \cite{briskorn2014aircraft, Salehipour2020}. FCFS technique sequences the aircraft for arrival based on the arrival aircraft's entry to the radar range. As the aircraft enters the predetermined radar window, the ATC allocates a scheduled time window for an aircraft to land. 

The problem's complexity stems from multiple interacting constraints and objectives: wake vortex separation requirements, time window restrictions, and the need to minimize both delay costs and fuel consumption \cite{Hammouri2019}. Given the size and complexity of the problem that requires computing a schedule of arriving aircraft on runways, researchers have tried to optimize delays and costs by implementing different solution approaches. \cite{Beasley2000} proposed a Mixed-Integer Programming (MIP) approach for the ALP and included different penalties for early and late landings of the aircraft than the scheduled landing time. They have managed a standardized OR-Library of the problem set at \url{http://people.brunel.ac.uk/~mastjjb/jeb/orlib/airlandinfo.html}. Improving upon the static case, the authors further expanded the MIP formulation to include a dynamic case where they incorporated displacement constraints \cite{beasley2004displacement}. \cite{ernst1999heuristic} developed a simplex-based solution approach for the ALP. part from the MIP solution approach, there are other operations research-based solutions, heuristics and meta-heuristics solutions, and ensemble approaches consisting of different combinations of the above methods. Literature also consists of dynamic programming and learning-based algorithms for the ALP. An overview of these methods is provided in the section \ref{sec:background}.

In the ALP, various formulations consist of a few common constraints across the literature. Some of those constraints are time window constraints, and separation constraints given the category of an aircraft. The International Civil Aviation Organization (ICAO) has defined the vertical and longitudinal separation constraints \cite{doc20164444}. These constraints help maintain the aircraft's aerodynamic stability and enhance the overall safety of the operation. There is a Wake Vortex (WV) separation that clarifies the separation of different aircraft based on the categorical classification: Super Heavy (S), Heavy (H), Medium (M), and Light (L). S category is only for the Aircraft - Airbus A380-800. There are further suggestions and improvements over this existing classification by the Federal Aviation Administration (FAA) as there are multiple factors that account for separation requirements, e.g., weather conditions. The time window includes the earliest possible time and the latest possible time. Depending on the aircraft's actual landing time and its difference from the scheduled landing time, the penalties are calculated \cite{Beasley2000}. 

Heuristic and metaheuristic algorithms often do not provide results as optimal as those produced by operations research methods. However, their computational efficiency can be competitive, making them attractive for real-world applications. This paper aims to address the shortcomings of certain traditional approaches, such as scalability issues and the lack of generalization and consistency in solutions. Due to these limitations, many methods in the literature face challenges in modern industrial environments. For example, given the dynamic nature of aircraft arrivals, real-time rescheduling is essential. If an algorithm requires significant computational time to update the schedule—even if it produces optimal results—its practical usefulness may be limited. This concern may explain why the FCFS approach is still in use today. 

The important contribution of this paper is a new way of approaching the ALP using Deep Reinforcement Learning (DRL). Given the recent successes of DRL in DeepMind's AlphaGo, AlphaZero, AlphaStar, and OpenAI's O1 model, there is an increased interest in assessing and observing new algorithms and their results of problems such as ALP. This paper uses an Actor-Critic DRL architecture combining graph neural networks (GNNs) with policy optimization for modeling the relationship structures between the aircraft and the constraints it should follow. This architecture helps in capturing both spatial and temporal aspects of runway scheduling. The architecture focuses on a time-varying perturbation mechanism that adaptively balances exploration and exploitation throughout the learning process to ensure efficient utilization of the runway while maintaining separation constraints. The proposed solution is tested on a standardized dataset for comprehensive comparison and evaluation. Detailed formulation and algorithm are provided in section \ref{sec:methodology} of this paper.

The next sections of the paper are organized as follows. Section \ref{sec:background} explains the ALP in detail and the differences between the standardized dataset and the particular instances of the dataset used in this paper. This section also provides an overview of existing methods and literature. In section \ref{sec:methodology}, the DRL formulation is provided. Which is followed by section \ref{sec:experiments} that discusses the results of algorithm's performance on standardized dataset and it's comparisons with different algorithms on selected instances. Section \ref{sec:conclusion} offers the conclusion and future directions of the research problem and potential solutions.

\section{Background and Problem Definition}\label{sec:background}
This section will address the ALP problem definition, data instances, and existing approaches from various fields. 

\subsection{Problem Definition}
The Aircraft Landing Problem (ALP) represents a critical operational challenge in air traffic management. At its core, ALP involves determining optimal landing times for arriving aircraft while satisfying multiple operational constraints and minimizing various cost factors \cite{Ikli2021}. Consider a set of aircraft $\mathcal{A} = \{1,2,...,n\}$ to be scheduled for landing on a runway. For each aircraft $i \in \mathcal{A}$:

\begin{itemize}
    \item $T_i$: Target landing time
    \item $E_i$: Earliest possible landing time
    \item $L_i$: Latest possible landing time
    \item $c^+_i$: Penalty cost per unit time for landing after $T_i$
    \item $c^-_i$: Penalty cost per unit time for landing before $T_i$
    \item $w_i$: Wake turbulence category (Heavy/Medium/Light)
\end{itemize}

For any pair of aircraft $(i,j)$ landing consecutively:
\begin{itemize}
    \item $s_{ij}$: Minimum separation time based on wake vortex categories
    \item $\delta_{ij}$: Binary variable indicating if i lands before j
    \item $\alpha_i$: Time units landed before target time
    \item $\beta_i$: Time units landed after target time
\end{itemize}

The primary decision variable for each aircraft $i$ is:
\begin{itemize}
    \item $x_i$: Actual landing time $x_i \in [E_i, L_i]$
\end{itemize}

\subsubsection{Objective Function}
Minimize the total weighted deviation from target times:

\begin{equation}
\min \sum_{i=1}^{n} (c^-_i\alpha_i + c^+_i\beta_i)
\end{equation}

\subsubsection{Constraints}

\begin{equation}
E_i \leq x_i \leq L_i, \quad \forall i \in \mathcal{A}
\end{equation}

\begin{equation}
\delta_{ij} + \delta_{ji} = 1, \quad \forall i,j \in \mathcal{A}, i \neq j
\end{equation}

\begin{equation}
x_j - x_i \geq s_{ij}\delta_{ij}, \quad \forall i,j \in \mathcal{A}, i \neq j
\end{equation}

\begin{equation}
x_i - T_i = \beta_i - \alpha_i, \quad \forall i \in \mathcal{A}
\end{equation}

\begin{equation}
0 \leq \alpha_i \leq T_i - E_i, \quad \forall i \in \mathcal{A}
\end{equation}

\begin{equation}
0 \leq \beta_i \leq L_i - T_i, \quad \forall i \in \mathcal{A}
\end{equation}

\begin{equation}
\delta_{ij} = 1, \quad \forall (i,j) \in \mathcal{P}
\end{equation}

\begin{equation}
x_j - x_i \geq s_{ij} + b, \quad \forall i,j \in \mathcal{A}, i \neq j, \delta_{ij} = 1
\end{equation}

\begin{equation}
x_i \geq 0, \quad \forall i \in \mathcal{A} \\
\end{equation}

\begin{equation}
\alpha_i, \beta_i \geq 0, \quad \forall i \in \mathcal{A} \\
\end{equation}

\begin{equation}
\delta_{ij} \in \{0,1\}, \quad \forall i,j \in \mathcal{A}, i \neq j
\end{equation}

The separation times $s_{ij}$ are defined according to ICAO standards based on the wake turbulence categories of the leading and following aircraft. Table \ref{tab:separation_requirements} shows different combinations of separation times between leading and following aircraft provided in \cite{Ikli2021}. For example, if aircraft $i$ is Heavy and aircraft $j$ is Light, then $s_{ij} = 240$ seconds. For any pair of aircraft $(i,j)$ landing consecutively, a minimum separation time $s_{ij}$ must be maintained based on aircraft weight categories (Heavy/Medium/Light) and landing sequence (leading vs following aircraft). The primary decision variable for each aircraft $i$ is actual landing time $t_i \in [E_i, L_i]$. The objective is to minimize the total cost of deviations from target times as provided in the objective function. 

Objetive funciton is constrained to certain operational requirements and safety consideration. The time window constraint (2) ensures each aircraft lands within its operational limits. The landing sequence constraints (3) establish the binary relationship between any pair of aircraft. For each pair $(i,j)$, either aircraft $i$ lands before $j$ ($\delta_{ij} = 1$) or $j$ lands before $i$ ($\delta_{ji} = 1$). This constraint creates a well-defined ordering of landings, essential for maintaining separation requirements and operational control \cite{Bennell2011}. The wake vortex separation constraint (4) is critical for safety. It enforces minimum time separation between consecutive landings based on the wake turbulence categories of the aircraft involved. These separations are mandated by aviation authorities and are non-negotiable safety requirements. The values of $s_{ij}$ vary according to ICAO standards, with larger separations required when a smaller aircraft follows a larger one \cite{Ikli2021}. Different combinations of separation is provided in Table \ref{tab:separation_requirements}. 

Constraints (5)-(6) model the deviation from target times, which is central to the cost calculation. The deviation is categorized into earliness ($\alpha_i$) and lateness ($\beta_i$) parameters, allowing different penalty costs to be applied. This separation is particularly important, as early and late arrivals often have different operational and economic implications \cite{Salehipour2013}. The standardized data used to test the proposed solution consist of delay cost penalties, provided in Table \ref{tab:cost_structure}. Constraint (7) is a precedence constraint. For ordered pairs of aircraft $(i,j) \in \mathcal{P}$ where $\mathcal{P}$ is the set of precedence relationships. The safety buffer constraint (8) adds an additional layer of protection beyond the minimum separation requirements, where $b$ is the minimum safety buffer time. This buffer $b$ helps absorb minor variations in approach speeds and compensates for weather conditions or other operational uncertainties. It represents a practical enhancement to the theoretical minimum separations \cite{Hammouri2019}. In proposed solution, these can be modified to any desired value during testing and implementation. Lastly, the non-negativity and binary constraints (9) ensure solution's mathematical feasibility. The formulation captures the essential elements of the ALP. Separation of penalties into early and late arrival allows for more nuanced cost modeling, reflecting the reality that early arrivals may be less costly than delays \cite{Beasley2000, Ikli2021}.

\subsection{Standardized Data and Instances}\label{subsec:data_instances}

This subsection discusses standardized datasets in the ALP, emphasizing their evolution in capturing operational complexity. This dataset standardization provides researchers with a robust foundation by enabling objective comparisons of algorithmic performance  – a crucial step toward real-world implementation. This comparison is useful for series of downstream decisions in aircraft operations, such as the required memory space, expected runway throughput, etc. The OR-Library \cite{Beasley2000} dominated early ALP research with synthetic instances featuring homogeneous aircraft distributions (3 wake turbulence categories), static separation matrices independent of ICAO regulations, linear penalty structures for time deviations, fixed 30-minute time windows ($[T_i - 15, T_i + 15]$), and scalability up to 500 aircraft for theoretical analysis. OR-Library instances of the ALP can be found here \url{https://people.brunel.ac.uk/~mastjjb/jeb/orlib/airlandinfo.html}.

OR-Library instances provided a standardized test for ALP algorithms for a considerable time. However, while enabling foundational algorithmic comparisons, OR-Library instances exhibit critical operational shortcomings. Some of those limitations include cost models lacking tiered delay penalties, absence of real-world traffic patterns, separation time inconsistencies involving ICAO/EUROCONTROL standards, etc. Addressing these limitations, to resemble real-life scenarios,  \cite{Ikli2021} introduced subset of Paris-Orly Airport (LFPO) operations from \cite{OpenSky2013} as a standardized dataset, featuring tiered delay penalties \ref{tab:cost_structure}, and ICAO-compliant separations \ref{tab:separation_requirements}. Their data includes certain patterns of peak traffic hours at the Paris-Orly airport (LFPO). Data can be found at \url{https://data.recherche.enac.fr/ikli-alp/}. The dataset incorporates ICAO wake turbulence categories with realistic distributions (\ref{tab:separation_requirements}, in their paper \cite{Ikli2021}, they provide detailed calculations of separation times between the leading and following aircraft. The data satisfies two critical components of this research: real-life data, and standardized to compare our algorithmmic performance.

\begin{table}[ht]
\caption{Cost Structure in Ikli Dataset}
\label{tab:cost_structure}
\centering
\begin{tabular}{ccc}
\toprule
Time Period (sec) & Cost Coefficient\\
\midrule
0-300 & cost\_300\\
301-900 & cost\_900 \\
901-1800 & cost\_1800\\
1800-3600 & cost\_3600\\
\bottomrule
\end{tabular}
\end{table}

\begin{table}[ht]
\caption{Separation Requirements (seconds)}
\label{tab:separation_requirements}
\centering
\begin{tabular}{llll}
\toprule
Leading/Following & Heavy & Medium & Light \\
\midrule
Heavy & 96 & 157 & 240 \\
Medium & 60 & 69 & 156 \\
Light & 60 & 69 & 82 \\
\bottomrule
\end{tabular}
\end{table}

Each instance in the \cite{Ikli2021} dataset contains unique identifier (sr), aircraft model (mdl), wake turbulence category, scheduled arrival time (sta), actual arrival time (ata), four-tier cost coefficients. This enhanced dataset provides a more challenging and realistic test for evaluating the ALP scheduling algorithms, particularly for assessing their practical applicability in operational environments. 

\subsection{Existing Solution Approaches and Research Landscape}\label{subsec:existing_approaches}
The aircraft landing problem has attracted significant research attention across multiple disciplines, resulting in diverse methodological approaches. This section provides a critical analysis of four major research directions, examining their technical contributions and operational limitations.
\subsubsection{Operations Research Approaches}
The foundational work by \cite{Beasley2000} established the modern formulation of ALP through mixed-integer programming (MIP), introducing distinct early/late penalty costs and creating the OR-Library benchmark dataset. Their static formulation considered single-runway scenarios with fixed separation matrices, demonstrating optimal solutions for instances up to 50 aircraft using branch-and-bound methods. The authors later extended this to dynamic scenarios \cite{beasley2004displacement} through displacement constraints that allowed limited schedule adjustments, though at the cost of quadratic growth in binary variables. \cite{ernst1999heuristic} developed a simplex-based solution approach that laid groundwork for linear programming relaxations in ALP.

Subsequent noteworthy operations research solutions addressed specific operational constraints. For example, \cite{Faye2015} developed time-indexed formulations with valid inequalities that improved solution quality for medium-sized instances. \cite{Avella2017} proposed column generation techniques for multi-runway systems, demonstrating improved computational efficiency. Furthermore, \cite{Ghoniem2015} introduced symmetry-breaking constraints that enhanced the performance of CPLEX for larger instances. While these exact methods guarantee optimality, their computational complexity limits practical application to real-time scenarios. Recent benchmarks show CPLEX requires significant time for large instances, making pure MIP approaches unsuitable for dynamic environments and real-time use cases.

\subsubsection{Heuristic and Metaheuristic Innovations}
Metaheuristic approaches emerged to address scalability challenges, particularly for large-scale instances. \cite{Salehipour2013} pioneered a hybrid Simulated Annealing-MIP approach combining local search with exact neighborhood evaluation. Their variable neighborhood descent mechanism showed improvements in solution quality while maintaining computational efficiency compared to pure MIP. \cite{Hammouri2019} later enhanced this approach with iterated local search components.
\cite{Pinol2006} developed a population-based scatter search method with adaptive reference sets, which incorporates path relinking to maintain solution diversity. Their bionomic algorithm variant introduced ecological selection pressures, showing promise for larger instances. \cite{Vadlamani2014} proposed novel hybrid heuristics specifically for single-runway scenarios. Using iterated local search \cite{Sabar2015} achieved promising results through dynamic perturbation strength adaptation. Their self-tuning mechanism alternated between insertion and swap operators based on solution landscape characteristics, demonstrating improvements over static ILS implementations. \cite{soykan2016tabu} developed a tabu search methodology with multiple runway constraints, demonstrating effective prevention of cycling through memory structures while achieving competitive solution quality for instances up to 25 aircraft.
Despite these advances, metaheuristics face reproducibility challenges -- \cite{Hammouri2019} showed solution quality variance across random seeds for identical instances. This stochasticity complicates deployment in safety-critical aviation systems.

\subsubsection{Dynamic and Constraint Programming Approaches}
Real-time scheduling requirements motivated temporal decomposition strategies in the ALP. \cite{Furini2015} implemented a sliding window approach with look-ahead horizons in their rolling horizon optimization. Their aircraft grouping strategy showed potential for reducing problem dimensionality while maintaining constraint satisfaction rates. \cite{Furini2012} extended this work with adaptive horizon adjustments.
\cite{Balakrishnan2010} approached ALP with Constrained Position Shifting (CPS). They developed dynamic programming with CPS $\leq$ 3, enabling polynomial-time solutions for FAA-compliant scenarios. Later extensions \cite{Lieder2015} incorporated aircraft class-specific shifting limits, showing potential for improved runway utilization.
\cite{Artiouchine2008} formulated runway sequencing as a cumulative scheduling problem with holding pattern constraints. Their edge-finding propagators showed promise for separation compliance but required airport-specific tuning of search strategies. \cite{VanLeeuwen2003} proposed constraint relaxation techniques that improved computational efficiency for large instances.
These methods excel in specific operational contexts but struggle with generalizability across different airport configurations and traffic patterns.

\subsubsection{Machine Learning Methodologies}
As the aircraft planning involves varieties of problems at different levels (strategic, tactical, and operational) \cite{lohatepanont2004airline}, certain noteworthy machine learning approaches are discussed here.
Recent machine learning approaches attempt to balance optimality with computational efficiency. \cite{BritoSoares2015} used Q-Learning for Departure Management, and demonstrated potential for reducing taxi times using discrete state-action spaces. However, their tabular representation faced scalability challenges for larger scenarios. \cite{Brittain2018} pioneered Deep Reinforcement Learning (DRL) for aircraft sequencing using convolutional networks on spatial-temporal grids. Their approach showed promise in decision times but faced challenges in constraint satisfaction.
\cite{Girish2016} combined particle swarm optimization with neural cost predictors in their hybrid neuro-symbolic methods, showing potential for reducing fuel consumption through aerodynamic wake modeling. \cite{Shohel2018} developed specialized genetic operators for runway configuration optimization. \cite{Bennell2017} introduced preference models balancing airline costs and ATC workload in their multi-stakeholder optimization, while \cite{Pohl2020} addressed winter operations through integrated de-icing scheduling. While promising, these early learning approaches either sacrificed constraint guarantees or required extensive manual engineering for state representation. Recent advances in neural architectures have shown promise for complex air traffic control tasks. \cite{guo2024integrating} demonstrated transformer networks' effectiveness in interpreting spoken ATC instructions for flight trajectory prediction, achieving 92.3\% accuracy in instruction parsing through their dual-encoder architecture. Their work establishes transformers' ability to process sequential ATC communications while maintaining temporal relationships in aircraft trajectories, a critical requirement for landing scheduling system. GNNs have emerged as powerful tools for industrial scheduling challenges. \cite{song2022flexible} recently applied GNNs to semiconductor manufacturing scheduling, achieving 18.7\% cycle time reduction through their hierarchical message passing architecture. Their work confirms GNNs' ability to capture complex resource allocation patterns that are useful in solving the ALP.

\section{Methodology}\label{sec:methodology}

The proposed methodology addresses the aircraft landing problem through a novel combination of deep reinforcement learning and graph neural networks. This integration allows us to capture both the temporal aspects of aircraft scheduling and the complex relationships between aircraft characteristics, separation requirements, and runway constraints. The approach differs from traditional optimization methods by learning scheduling policies that can adapt to varying conditions while maintaining computational efficiency.

\subsection{Deep Reinforcement Learning Framework}
This paper proposes an ALP solution as a Markov Decision Process (MDP) that is solved using a graph-enhanced deep reinforcement learning approach. The framework combines graph neural networks with actor-critic architecture. The effectiveness of reinforcement learning depends heavily on how we represent the problem state and possible actions. In this formulation, spaces are carefully designed to capture all relevant operational factors while maintaining tractability. The state space incorporates both static aircraft characteristics and dynamic scheduling conditions, while the action space is structured to ensure feasible landing time assignments.

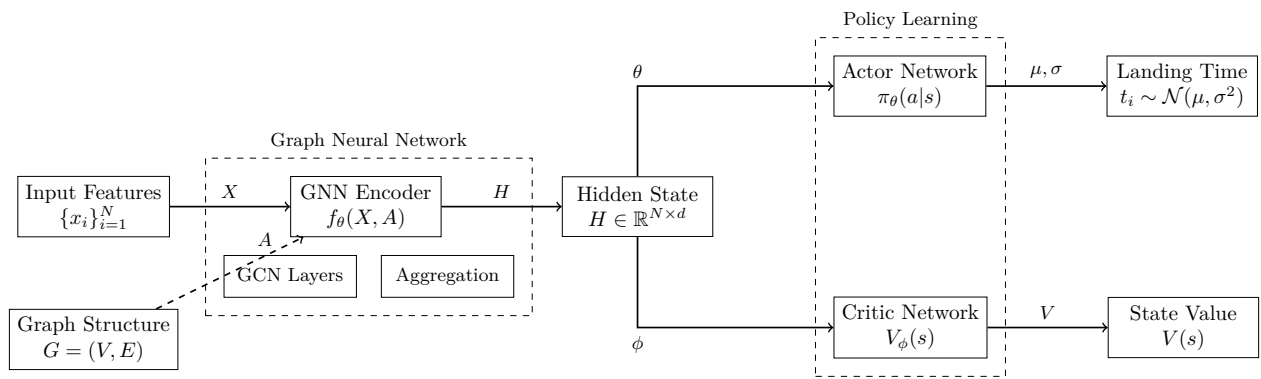
\begin{figure}[ht]
    \centering
    \begin{tikzpicture}[
    scale=1.0,  
    node distance=2.5cm,  
    box/.style={rectangle,draw,minimum width=2.5cm,minimum height=1cm,font=\small,align=center},
    feature/.style={rectangle,draw,minimum width=2.2cm,minimum height=0.7cm,font=\footnotesize},  
    arrow/.style={->,thick},
    doublearrow/.style={<->,thick}
]
\node[box] (input) {Input Features\\$\{x_i\}_{i=1}^N$};
\node[box, below=1.2cm of input] (graph) {Graph Structure\\$G=(V,E)$};

\node[box, right=2cm of input] (gnn) {GNN Encoder\\$f_{\theta}(X,A)$};
\node[feature, below=0.3cm of gnn.south west] (gnn1) {GCN Layers};
\node[feature, right=0.4cm of gnn1] (gnn2) {Aggregation};

\node[box, right=2cm of gnn] (hidden) {Hidden State\\$H \in \mathbb{R}^{N \times d}$};

\node[box, above right=1cm and 2cm of hidden] (actor) {Actor Network\\$\pi_{\theta}(a|s)$};
\node[box, below right=1cm and 2cm of hidden] (critic) {Critic Network\\$V_{\phi}(s)$};

\node[box, right=2cm of actor] (action) {Landing Time\\$t_i \sim \mathcal{N}(\mu,\sigma^2)$};
\node[box, right=2cm of critic] (value) {State Value\\$V(s)$};

\draw[arrow] (input) -- node[above,font=\footnotesize] {$X$} (gnn);
\draw[arrow, dashed] (graph) -- node[above right=0.45cm,font=\footnotesize] {$A$} (gnn);
\draw[arrow] (gnn) -- node[above,font=\footnotesize] {$H$} (hidden);
\draw[arrow] (hidden) |- node[above,font=\footnotesize] {$\theta$} (actor);
\draw[arrow] (hidden) |- node[below,font=\footnotesize] {$\phi$} (critic);
\draw[arrow] (actor) -- node[above,font=\footnotesize] {$\mu,\sigma$} (action);
\draw[arrow] (critic) -- node[above,font=\footnotesize] {$V$} (value);

\begin{pgfonlayer}{background}
    \node[rectangle,draw,dashed,fit=(gnn) (gnn1) (gnn2),inner sep=0.3cm,
          label=above:{\footnotesize Graph Neural Network}] {};
    \node[rectangle,draw,dashed,fit=(actor) (critic),inner sep=0.3cm,
          label=above:{\footnotesize Policy Learning}] {};
\end{pgfonlayer}

\end{tikzpicture}
    \caption{Architecture of the proposed GNN-based Actor-Critic network for aircraft landing scheduling. The model combines graph neural network layers for feature extraction with parallel actor and critic heads for landing time prediction and value estimation.}
    \label{fig:model_architecture}
\end{figure}

\subsubsection{State Space}
The state space $s_t$ at time $t$ comprises a set of features for each aircraft $i \in \mathcal{A}$:
\begin{itemize}
    \item Normalized scheduled arrival time $\bar{t}^s_i \in [0,1]$
    \item Aircraft category encoding $c_i \in \{0,1,2\}$ for Heavy/Medium/Light
    \item Normalized urgency level $\bar{u}_i$ derived from cost coefficients
    \item Normalized landing time window $[\bar{E}_i, \bar{L}_i]$
    \item Priority score $p_i$ computed as:
    \begin{equation}
    p_i = \omega_1\bar{u}_i + \omega_2\tau_i + \omega_3\kappa_i + \omega_4\bar{c}_i 
    \end{equation}
    
    where:
    \begin{itemize}
        \item $\tau_i$ is time window criticality
        \item $\kappa_i$ is category-based priority
        \item $\bar{c}_i$ is normalized cost factor
        \item $\omega_1, ..., \omega_4$ are weighting coefficients
    \end{itemize}
\end{itemize}

\subsubsection{Action Space}
For each aircraft $i$, the action space defines:
\begin{equation}
a_i = t_i \in [E_i, L_i]
\end{equation}
where $t_i$ represents the assigned landing time within the feasible time window.

\subsubsection{Reward Function Design}\label{subsubsec:reward}
The reward function incorporates multiple components to guide learning toward optimal scheduling:

\begin{equation}
R(s,a) = w_1R_{delay} + w_2R_{separation} + w_3R_{throughput} + w_4R_{smoothness}
\end{equation}

where:
\begin{itemize}
\item $R_{delay}$ penalizes deviations from target times using the tiered cost structure
\item $R_{separation}$ provides positive rewards for maintaining required separation
\item $R_{throughput}$ encourages high runway utilization
\item $R_{smoothness}$ penalizes rapid changes in landing intervals
\end{itemize}

The weights $w_1,\ldots,w_4$ are tuned to balance these objectives, with values:
\begin{equation}\label{eq:weights}
[w_1,w_2,w_3,w_4] = [1.0, 2.0, 0.5, 0.3]
\end{equation}

\subsection{Main Algorithm}
The overall learning procedure is outlined in Algorithm \ref{alg:main_drl}, which coordinates the interaction between the actor-critic network and the environment:

\begin{algorithm}[H]
\caption{Main Deep Reinforcement Learning Algorithm for ALP} 
\label{alg:main_drl}
\begin{algorithmic}[1]
\REQUIRE Aircraft data, network parameters, hyperparameters
\STATE Initialize actor-critic networks with parameters $\theta$, $\phi$
\STATE Initialize environment with aircraft time windows and constraints
\WHILE{training not converged}
    \STATE Reset environment
    \STATE $s_0 \leftarrow$ initial state
    \STATE $done \leftarrow$ False
    \WHILE{not done}
        \STATE Construct graph $G$ from current state
        \STATE $a_t \leftarrow$ Actor($s_t|\theta$) with exploration $\epsilon_t$
        \STATE $t_i \leftarrow$ Safety-Aware-Landing-Assignment($a_t$)
        \STATE $s_{t+1}$, $r_t$, $done \leftarrow$ Environment($s_t$, $t_i$)
        \STATE Store transition $(s_t, a_t, r_t, s_{t+1})$ in buffer
        \STATE Update $\theta$, $\phi$ using stored transitions
        \STATE $s_t \leftarrow s_{t+1}$
    \ENDWHILE
\ENDWHILE
\end{algorithmic}
\end{algorithm}

The environment step function calculates rewards based on the tiered cost structure:

\begin{algorithm}[H]
\caption{Environment Step Function}
\begin{algorithmic}[1]
\REQUIRE Current state $s_t$, Action $t_i$
\STATE Calculate delay $d_i = |t_i - T_i|$
\STATE Initialize cost = 0
\IF{$d_i \leq 300$}
    \STATE cost = $d_i \times$ cost\_300
\ELSIF{$d_i \leq 900$}
    \STATE cost = 300 $\times$ cost\_300 + $(d_i-300) \times$ cost\_900 
\ELSIF{$d_i \leq 1800$}
    \STATE cost = 300 $\times$ cost\_300 + 600 $\times$ cost\_900 + $(d_i-900) \times$ cost\_1800
\ELSE
    \STATE cost = 300 $\times$ cost\_300 + 600 $\times$ cost\_900 + 900 $\times$ cost\_1800 + $(d_i-1800) \times$ cost\_3600
\ENDIF
\STATE reward = -cost + safety\_bonus
\STATE Update aircraft landing schedule
\RETURN next\_state, reward, done
\end{algorithmic}
\end{algorithm}

\subsection{Graph Construction and Priority Rules}

Graph representation provides an intuitive and powerful way to model relationships between aircraft in the landing sequence. By representing each aircraft as a node and their scheduling dependencies as edges, we can leverage graph neural networks to learn complex patterns in aircraft sequencing. This graph-based approach naturally captures separation requirements and allows the model to reason about global scheduling constraints.

\subsubsection{Graph Structure}
For each state $s_t$, a fully connected directed graph $G = (V,E)$ where:

\begin{itemize}
    \item Vertices $V$ represent aircraft, with $|V| = n$ for $n$ aircraft
    \item Edges $E$ represent potential landing sequence dependencies
    \item Node features matrix $X \in \mathbb{R}^{|V| \times d}$ where $d$ is feature dimension
    \item Adjacency matrix $A \in \mathbb{R}^{|V| \times |V|}$ representing landing sequence relationships
\end{itemize}

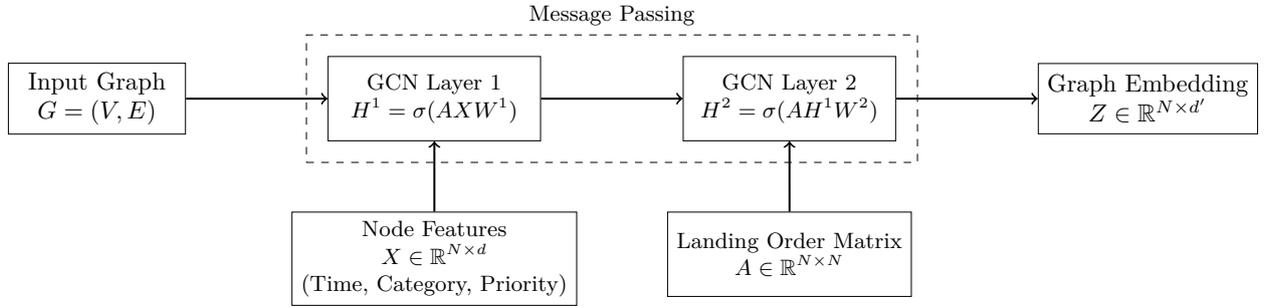
\begin{figure}[ht]
    \centering
    \begin{tikzpicture}[
    scale=0.8,
    node distance=2cm,
    box/.style={
        rectangle,
        draw,
        minimum width=2.5cm,
        minimum height=1cm,
        font=\small,
        align=center
    },
    layer/.style={
        rectangle,
        draw,
        minimum width=3cm,
        minimum height=1.2cm,
        font=\footnotesize,
        align=center  
    },
    arrow/.style={->,thick}
]

\node[box] (input) {Input Graph\\$G=(V,E)$};

\node[layer, right=2cm of input] (gcn1) {
    GCN Layer 1\\
    $H^1 = \sigma(AXW^1)$
};
\node[layer, right=2cm of gcn1] (gcn2) {
    GCN Layer 2\\
    $H^2 = \sigma(AH^1W^2)$
};

\node[layer, below=1cm of gcn1] (feat) {
    Node Features\\
    $X \in \mathbb{R}^{N \times d}$\\
    (Time, Category, Priority)
};
\node[layer, below=1cm of gcn2] (adj) {
    Landing Order Matrix\\
    $A \in \mathbb{R}^{N \times N}$
};

\node[box, right=2cm of gcn2] (output) {Graph Embedding\\$Z \in \mathbb{R}^{N \times d'}$};

\draw[arrow] (input) -- (gcn1);
\draw[arrow] (gcn1) -- (gcn2);
\draw[arrow] (gcn2) -- (output);
\draw[arrow] (feat) -- (gcn1);
\draw[arrow] (adj) -- (gcn2);

\node[rectangle,draw,dashed,fit=(gcn1)(gcn2),  
      inner sep=0.3cm,
      label={[above,font=\footnotesize]Message Passing}] {};

\end{tikzpicture}
    \caption{Graph neural network architecture showing message passing between aircraft nodes. The GNN encoder transforms raw aircraft features into a learned representation that captures both spatial and temporal relationships.}
    \label{fig:gnn_arch}
\end{figure}

\subsubsection{Node Feature Construction}
Each node $i \in V$ contains the following features:
\begin{equation}
x_i = [\bar{t}^s_i, c_i, \bar{u}_i, \bar{E}_i, \bar{L}_i, p_i, \bar{l}_i]
\end{equation}

where:
\begin{itemize}
    \item $\bar{t}^s_i$: Normalized scheduled time $\in [0,1]$
    \item $c_i$: One-hot encoded aircraft category
    \item $\bar{u}_i$: Normalized urgency
    \item $[\bar{E}_i, \bar{L}_i]$: Normalized time window
    \item $p_i$: Priority score
    \item $\bar{l}_i$: Normalized current landing time (if assigned)
\end{itemize}

\subsubsection{Priority Score Calculation}
The priority calculation procedure is outlined in Algorithm \ref{alg:priority}:

\begin{algorithm}[H]
\caption{Priority Score Calculation}
\label{alg:priority}
\begin{algorithmic}[1]
\REQUIRE Aircraft $i$, Cost coefficients, Time window
\STATE Calculate urgency $\bar{u}_i$ from cost coefficients
\STATE // 1500 represents 25 minutes in seconds for time window normalization
\STATE time\_criticality $\leftarrow 1 - \frac{L_i - E_i}{1500}$
\STATE category\_priority $\leftarrow 1 - \frac{c_i}{2}$
\STATE cost\_factor $\leftarrow$ normalize(mean(cost coefficients))
\STATE priority $\leftarrow w_1\bar{u}_i + w_2$ time\_criticality $+ w_3$ category\_priority $+ w_4$ cost\_factor
\RETURN priority
\end{algorithmic}
\end{algorithm}

The components are calculated as follows:

1. Time Window Criticality ($\tau_i$):
\begin{equation}
\tau_i = 1 - \frac{L_i - E_i}{1500}
\end{equation}

2. Category Priority ($\kappa_i$):
\begin{equation}
\kappa_i = 1 - \frac{c_i}{2}
\end{equation}

3. Urgency ($\bar{u}_i$):

\begin{equation}
\bar{u}_i= \frac{u_i - u_{min}}{u_{max} - u_{min}}
\end{equation}

4. Cost Factor ($\bar{c}_i$):
\begin{equation}
\bar{c}_i = \frac{1}{4}\sum_{k \in K}\frac{c_{ik}}{\max_{j \in \mathcal{A}}c_{jk}}
\end{equation}
where $K = \{300s, 900s, 1800s, 3600s\}$ represents the cost tiers. \cite{Ikli2021} shows the normally time windows range from 600-900 seconds before and after 'sta', respectively. Therefore, to incorporate criticality, the paper uses 1500 seconds of the time window size.

\subsubsection{Edge Weight Assignment}
For each edge $(i,j) \in E$, a weight $w_{ij}$ representing the feasibility of aircraft $j$ following aircraft $i$:

\begin{equation}
w_{ij} = f(s_{ij}, \Delta t_{ij}, p_i, p_j)
\end{equation}

where:
\begin{itemize}
    \item $s_{ij}$: Required separation time
    \item $\Delta t_{ij}$: Time difference between target times
    \item $f(\cdot)$: Weighting function combining separation and priority factors
\end{itemize}

\subsection{Graph Neural Network Architecture}

\subsubsection{Message Passing Layers}
The graph neural network processes node features through message passing operations:

\begin{equation}
m_i^l = \sum_{j \in \mathcal{N}(i)} \text{MLP}_\theta(h_i^{l-1} \| h_j^{l-1} \| e_{ij})
\end{equation}

\begin{equation}
h_i^l = \text{GRU}(h_i^{l-1}, m_i^l)
\end{equation} where $\|$ denotes feature concatenation, $e_{ij}$ represents edge features, and $\text{MLP}_\theta$ is a multi-layer perceptron with parameters $\theta$. The GRU update mechanism helps maintain temporal dependencies.

The attention mechanism employs scaled dot-product attention:

\begin{equation}
\alpha_{ij} = \frac{\exp(e_{ij}/\sqrt{d_k})}{\sum_{k \in \mathcal{N}(i)} \exp(e_{ik}/\sqrt{d_k})}
\end{equation}

These operations are implemented in two successive layers:

\begin{equation}
H^1 = \sigma(AXW^1), \quad H^2 = \sigma(AH^1W^2)
\end{equation} where $\sigma$ is the ReLU activation function and $W^1, W^2$ are learnable weight matrices.

\subsubsection{Graph Attention Mechanism}
Approach proposes a multi-head attention mechanism:

\begin{equation}
\alpha_{ij}^k = \frac{\exp(e_{ij}^k)}{\sum_{l \in \mathcal{N}_i} \exp(e_{il}^k)}
\end{equation} where $e_{ij}^k$ is the attention coefficient for head $k$ between nodes $i$ and $j$. The graph neural network architecture enables message passing between aircraft nodes, allowing the model to learn implicit relationships between aircraft characteristics, separation requirements, and runway constraints. This learned representation captures complex scheduling patterns that would be difficult to express in traditional optimization formulations.

\subsection{Actor-Critic Network}

The actor-critic architecture combines the advantages of both policy-based and value-based learning. The actor network learns to propose landing times that satisfy operational constraints, while the critic network evaluates these decisions to guide policy improvement. This dual network structure helps balance exploration of new scheduling strategies with exploitation of known good solutions.

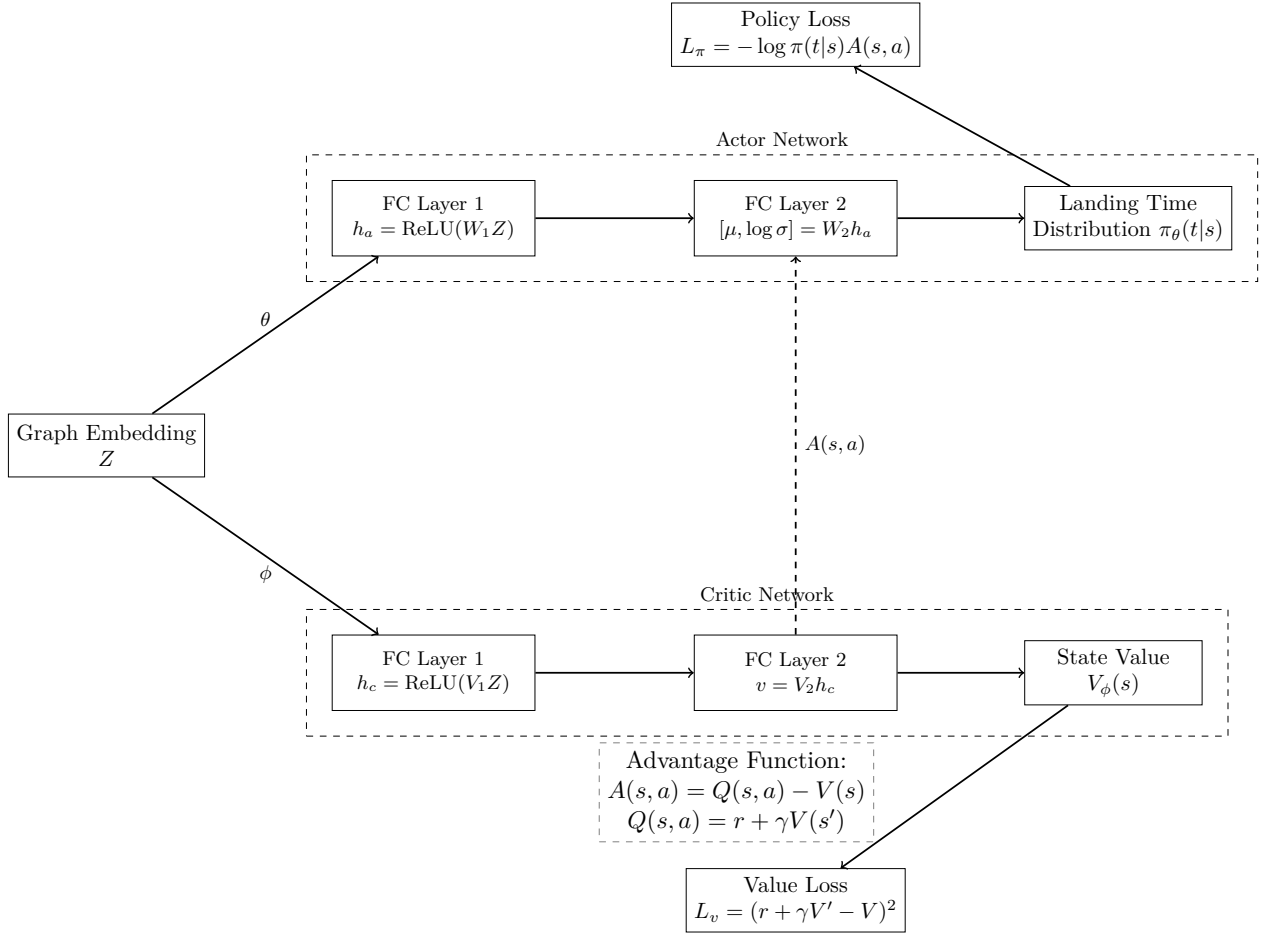
\begin{figure}[ht]
    \centering
    \begin{tikzpicture}[
    scale=0.8,
    node distance=2cm,
    box/.style={
        rectangle,
        draw,
        minimum width=2.8cm,
        minimum height=1cm,
        font=\small,
        align=center
    },
    net/.style={
        rectangle,
        draw,
        minimum width=3.2cm,
        minimum height=1.2cm,
        font=\footnotesize,
        align=center  
    },
    arrow/.style={->,thick}
]

\node[box] (input) {Graph Embedding\\$Z$};

\node[net, above right=2.5cm and 2cm of input] (actor1) {
    FC Layer 1\\
    $h_a = \text{ReLU}(W_1Z)$
};
\node[net, right=2.5cm of actor1] (actor2) {
    FC Layer 2\\
    $[\mu,\log\sigma] = W_2h_a$
};
\node[box, right=2cm of actor2] (policy) {
    Landing Time\\
    Distribution $\pi_\theta(t|s)$
};

\node[net, below right=2.5cm and 2cm of input] (critic1) {
    FC Layer 1\\
    $h_c = \text{ReLU}(V_1Z)$
};

\node[net, right=2.5cm of critic1] (critic2) {  
    FC Layer 2\\
    $v = V_2h_c$
};
\node[box, right=2cm of critic2] (value) {
    State Value\\
    $V_\phi(s)$
};

\node[box, below=2.5cm of critic2] (vloss) {
    Value Loss\\
    $L_v = (r + \gamma V' - V)^2$
};
\node[box, above=1.8cm of actor2] (ploss) {
    Policy Loss\\
    $L_\pi = -\log\pi(t|s)A(s,a)$
};

\node[draw=black!50, dashed, align=center, below right=0.5cm and 1cm of critic1] (advantage) {
    Advantage Function:\\
    $A(s,a) = Q(s,a) - V(s)$\\
    $Q(s,a) = r + \gamma V(s')$
};

\draw[arrow] (input) -- node[above,font=\footnotesize] {$\theta$} (actor1);
\draw[arrow] (actor1) -- (actor2);
\draw[arrow] (actor2) -- (policy);
\draw[arrow] (input) -- node[below,font=\footnotesize] {$\phi$} (critic1);
\draw[arrow] (critic1) -- (critic2);
\draw[arrow] (critic2) -- (value);
\draw[arrow] (policy) -- (ploss);
\draw[arrow] (value) -- (vloss);

\draw[arrow, dashed] (critic2) -- node[right,font=\footnotesize] {$A(s,a)$} (actor2);

\begin{pgfonlayer}{background}
    \node[rectangle,draw,dashed,fit=(actor1)(actor2)(policy),
          inner sep=0.4cm,
          label={[above,font=\footnotesize]Actor Network}] {};
    \node[rectangle,draw,dashed,fit=(critic1)(critic2)(value),
          inner sep=0.4cm,
          label={[above,font=\footnotesize]Critic Network}] {};
\end{pgfonlayer}

\end{tikzpicture}
    \caption{Actor-critic network architecture for learning landing policies. The actor network outputs landing time distributions while the critic estimates state values to guide policy improvement.}
    \label{fig:actor_critic}
\end{figure}

\subsubsection{Actor Network}
The actor network outputs landing time distribution parameters:

\begin{equation}
[\mu, \log\sigma] = f_\theta(Z)
\end{equation}

Landing time prediction is computed as:
\begin{equation}
t_i = E_i + \sigma(\mu_i)(L_i - E_i)
\end{equation}

The variance parameter $\sigma$ is clamped to ensure exploration stability:
\begin{equation}
\sigma = \exp(\text{clamp}(\log\sigma, -10, 2)) + \epsilon
\end{equation}
where $\epsilon = 10^{-5}$ is added for numerical stability.

\subsubsection{Critic Network}
State value estimation:
\begin{equation}
V(s) = g_\phi(Z)
\end{equation}

The architecture of both networks consists of fully connected layers with ReLU activations:

\begin{itemize}
    \item Actor: Input $\rightarrow$ FC(256) $\rightarrow$ ReLU $\rightarrow$ FC(128) $\rightarrow$ [$\mu$, $\log\sigma$]
    \item Critic: Input $\rightarrow$ FC(256) $\rightarrow$ ReLU $\rightarrow$ FC(128) $\rightarrow$ $V(s)$
\end{itemize}

\subsection{Policy Learning and Optimization}\label{subsec:policy_learning}
The policy learning process involves iterative optimization of both actor and critic networks through carefully structured loss functions and gradient updates. The policy gradient is computed using the advantage function:

\begin{equation}\label{eq:advantage}
A(s,a) = Q(s,a) - V_\phi(s)
\end{equation}

where $Q(s,a)$ represents the state-action value and $V_\phi(s)$ is the critic's value estimate.

The actor's loss function incorporates both the policy gradient and an entropy term to encourage exploration:

\begin{equation}\label{eq:actor_loss}
L_{actor} = -\mathbb{E}[\log \pi_\theta(a|s)A(s,a)] - \lambda H(\pi_\theta)
\end{equation}

where $\lambda$ is the entropy coefficient (set to 0.02 in implementation) and $H(\pi_\theta)$ is the policy entropy.

The critic is trained to minimize the temporal difference error:

\begin{equation}\label{eq:critic_loss}
L_{critic} = \mathbb{E}[(r_t + \gamma V_\phi(s_{t+1}) - V_\phi(s_t))^2]
\end{equation}

The implementation uses a time-varying learning rate schedule:

\begin{equation}\label{eq:learning_rate}
\alpha_t = \alpha_{end} + (\alpha_{start} - \alpha_{end})e^{-t/\tau}
\end{equation}

where $\alpha_{start} = 1e^{-4}$, $\alpha_{end} = 1e^{-5}$, and $\tau$ controls the decay rate based on training progress.

\subsubsection{Optimization Process}
The network parameters are updated using Adam optimizer with gradient clipping:

\begin{equation}
\theta_{t+1} = \theta_t - \alpha_t \cdot \text{clip}(\nabla_\theta L_{total}, -c, c)
\end{equation}

where $L_{total} = L_{actor} + L_{critic}$ and $c=10.0$ is the gradient clipping threshold. The learning rate $\alpha_t$ follows an exponential decay schedule to promote convergence while maintaining exploration capability in early training stages.

\subsection{Training Procedure}
The training process is designed to gradually improve scheduling performance while maintaining safety constraints. Through careful balancing of exploration and exploitation, the model learns to generate efficient schedules while respecting operational requirements. The time-varying exploration rate helps the model discover diverse scheduling strategies early in training while focusing on refinement in later stages.

\subsubsection{Loss Functions}
Actor and Critic losses are defined as:

\begin{equation}
L_{actor} = -\mathbb{E}[\log \pi_\theta(a|s)A(s,a)] - \lambda H(\pi_\theta)
\end{equation}

\begin{equation}
L_{critic} = \mathbb{E}[(R + \gamma V_\phi(s') - V_\phi(s))^2]
\end{equation}

where:
\begin{itemize}
    \item $A(s,a)$ is the advantage function: $A(s,a) = Q(s,a) - V_\phi(s)$
    \item $H(\pi_\theta)$ is the policy entropy
    \item $\lambda = 0.02$ is the entropy coefficient
    \item $\gamma = 0.99$ is the discount factor
\end{itemize}

\begin{algorithm}[H]
\caption{Training Procedure}
\label{alg:training}
\begin{algorithmic}[1]
\REQUIRE Batch of transitions $(s_t, a_t, r_t, s_{t+1})$
\STATE Calculate advantages $A_t = r_t + \gamma V_\phi(s_{t+1}) - V_\phi(s_t)$
\STATE Update critic by minimizing $L_{critic}$
\STATE Calculate policy gradient using $A_t$
\STATE Update actor by minimizing $L_{actor}$
\STATE Update exploration rate $\epsilon_t$
\end{algorithmic}
\end{algorithm}

\subsection{Adaptive Exploration Strategy}\label{subsec:exploration}
The framework employs a sophisticated exploration mechanism that combines $\epsilon$-greedy exploration with parameter noise. The exploration rate $\epsilon_t$ follows a time-varying schedule:

\begin{equation}\label{eq:epsilon}
\epsilon_t = \epsilon_{end} + (\epsilon_{start} - \epsilon_{end})e^{-t/\tau}
\end{equation} where $\epsilon_{start} = 0.9$ and $\epsilon_{end} = 0.3$. Additionally, Gaussian noise is added to the actor's parameters during exploration:

\begin{equation}\label{eq:param_noise}
\theta_{noisy} = \theta + \sigma_t\mathcal{N}(0,I)
\end{equation}

The noise scale $\sigma_t$ is adaptively adjusted based on the desired level of action perturbation:

\begin{equation}\label{eq:noise_scale}
\sigma_t = \sigma_{t-1}\cdot\alpha^{\text{sign}(d_{target} - d_t)}
\end{equation} where $d_t$ is the mean distance between perturbed and original actions, $d_{target}$ is the target distance, and $\alpha$ is the adaptation rate.

\subsection{Safety-Aware Sequential Assignment}

While deep learning provides powerful optimization capabilities, ensuring strict adherence to safety constraints is paramount in aviation applications. Proposed solution's safety-aware assignment process acts as a bridge between learned policies and operational requirements, guaranteeing that all proposed schedules maintain mandatory separation times and time window constraints. This hybrid approach combines the flexibility of learning-based methods with the reliability of rule-based systems.

\subsubsection{Assignment Algorithm}
The safety-aware sequential assignment process ensures constraint satisfaction:

\begin{algorithm}[H]
\caption{Safety-Aware Sequential Landing Assignment}
\label{alg:safety_assignment}
\begin{algorithmic}[1]
\REQUIRE Current state $s_t$, Predicted landing times $\{t_i\}$
\STATE Sort aircraft by priority score
\FOR{each aircraft $i$ in priority order}
    \STATE attempts $\leftarrow$ 0
    \WHILE{attempts $<$ max\_attempts}
        \IF{ValidateSeparation($t_i$)}
            \STATE Assign landing time $t_i$
            \STATE \textbf{break}
        \ELSE
            \STATE $t_i \leftarrow$ AdjustLandingTime($t_i$)
            \STATE attempts $\leftarrow$ attempts + 1
        \ENDIF
    \ENDWHILE
    \IF{attempts = max\_attempts}
        \STATE Apply recovery procedure
    \ENDIF
\ENDFOR
\RETURN Updated schedule
\end{algorithmic}
\end{algorithm}

\subsubsection{Separation Validation}
The separation validation ensures safety constraints:

\begin{algorithm}[H]
\caption{Validate Separation}
\label{alg:validate_separation}
\begin{algorithmic}[1]
\REQUIRE Aircraft $i$, Landing time $t_i$, Schedule
\FOR{each scheduled aircraft $j$}
    \STATE $\Delta t \leftarrow |t_i - t_j|$
    \STATE $s_{req} \leftarrow$ RequiredSeparation($i$, $j$)
    \IF{$\Delta t < s_{req} + \text{buffer}$}
        \RETURN False
    \ENDIF
\ENDFOR
\RETURN True
\end{algorithmic}
\end{algorithm}

The separation validation process includes a buffer time $b$ for enhanced safety:

\begin{equation}
\Delta t_{ij} \geq s_{ij} + b, \quad \forall i,j \in \mathcal{A}, i \neq j
\end{equation}

When violations occur, the algorithm iteratively adjusts within $[E_i, L_i]$ and attempts 7 times per aircraft (this number can be modified to satisfy computational efficiency) to find a suitable arrival time. In given attempts, if there is no feasible arrival time satisfying all the constraints, the model backtracks to the last feasible state. For enhanced security and weather-affected airports, the proposed solution can include buffer time as shown in the above equation, e.g., where $b$ is typically set to 30 seconds in implementation (on top of the separation time). These changes can be instantaneous. This methodology combines the representation power of graph neural networks with the decision-making capabilities of reinforcement learning while maintaining operational constraints and safety requirements. The multi-component architecture ensures both learning efficiency and operational safety in aircraft landing scheduling.

\section{Experimental Results and Analysis}\label{sec:experiments}

The proposed solution is tested on the Paris-Orly Airport dataset introduced by \cite{Ikli2021}. The dataset encompasses four distinct time periods representing different traffic patterns throughout the day: morning peak (07:00-11:00), mid-day (11:00-15:00), afternoon peak (15:00-19:00), and evening (19:00-23:00). This diversity in problem instances allows to evaluate this approach under varying operational conditions and traffic intensities. All experiments were performed on a single machine (M3 max processor). The implementation was done in Python.

\subsection{Training Convergence Analysis}

The training process evaluates the model across different scenario sizes, with systematic progression from smaller to larger aircraft sets. Scenario sizes were configured beginning with 5 aircraft first, and then progressing with 10 aircraft data size. Third and fourth scenarios had 20, and 30 aircraft, respectively. The training process systematically evaluates the model across increasing scenario complexities, progressing from 5 aircraft to 30 aircraft, with 4 meta-iterations (scenarios) and 10,000 episodes per scenario. At each stage, the learning process begins by initializing the actor-critic networks and environment with the corresponding scenario data. The exploration strategy employs a time-varying epsilon parameter, starting at 0.9 and decaying to 0.3, which controls the balance between exploring new solutions and exploiting learned policies. During each episode, the environment is reset and the network processes the current state through a graph-based representation capturing aircraft relationships and constraints. For action selection, the model either explores random feasible landing times or exploits the current policy based on the epsilon value. A safety-aware landing assignment mechanism validates all scheduling decisions against operational constraints, including separation requirements and time windows. The network parameters are updated using the Adam optimizer with a learning rate of $1e^{-4}$ and gradient clipping at 10.0 to ensure stable training. The model learns a complex scheduling pattern through progressive training and enforcing safety constraints. 

\begin{figure}[hbt!]
    \centering
    \subfloat[Scenario 1 Delays]{{\includegraphics[width=0.45\textwidth]{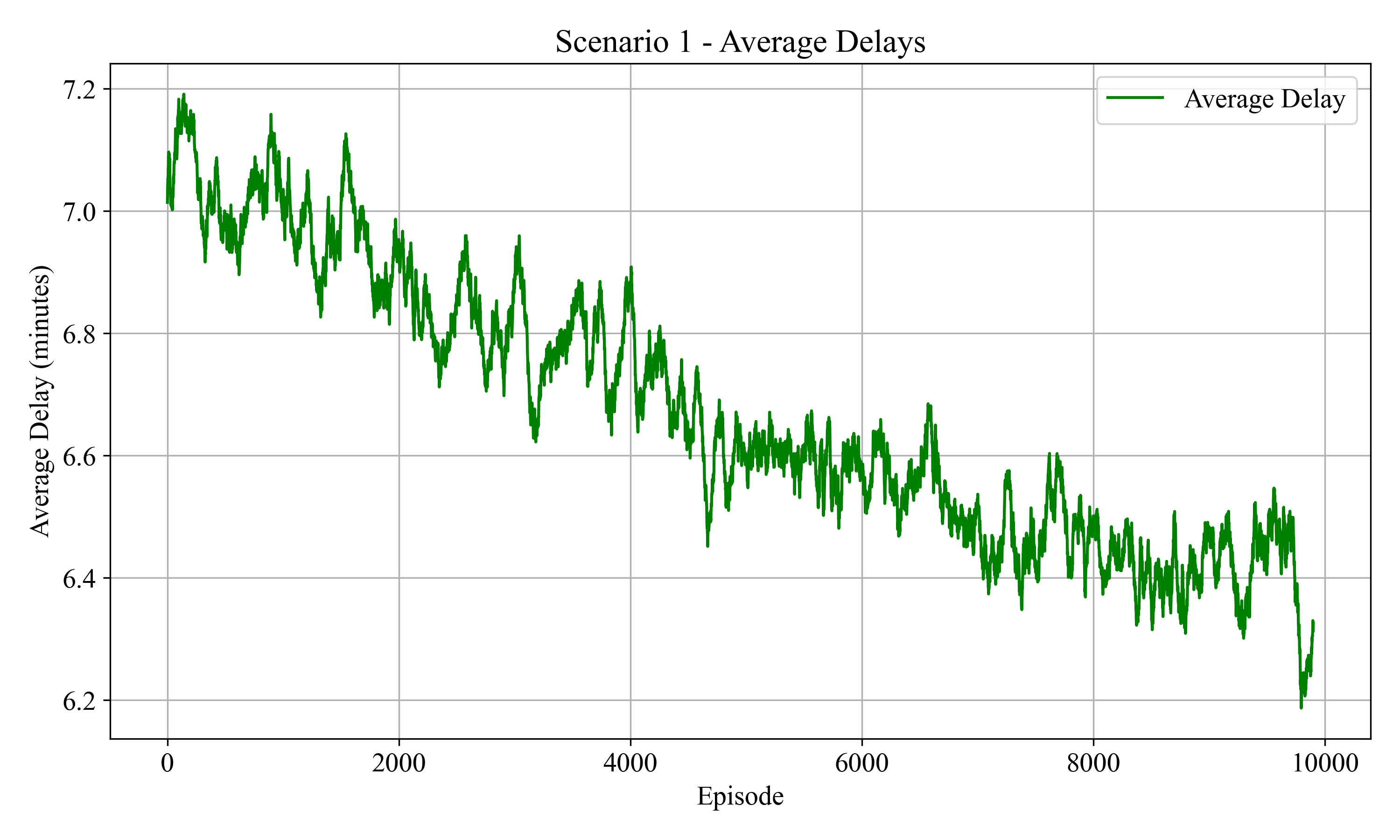} }}%
    \quad
    \subfloat[Scenario 2 Delays]{{\includegraphics[width=0.45\textwidth]{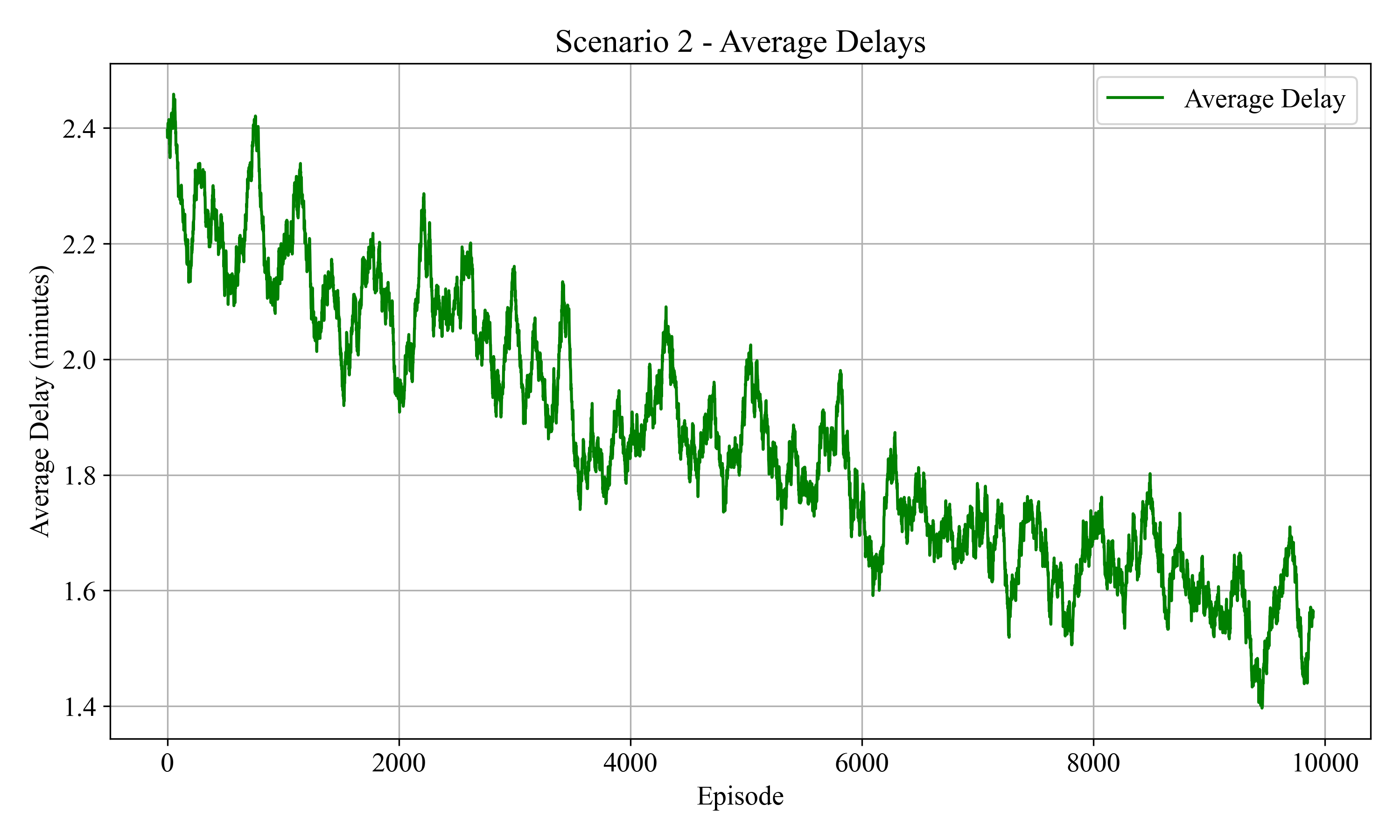} }}%
    
    \vspace{1em}
    
    \subfloat[Scenario 3 Delays]{{\includegraphics[width=0.45\textwidth]{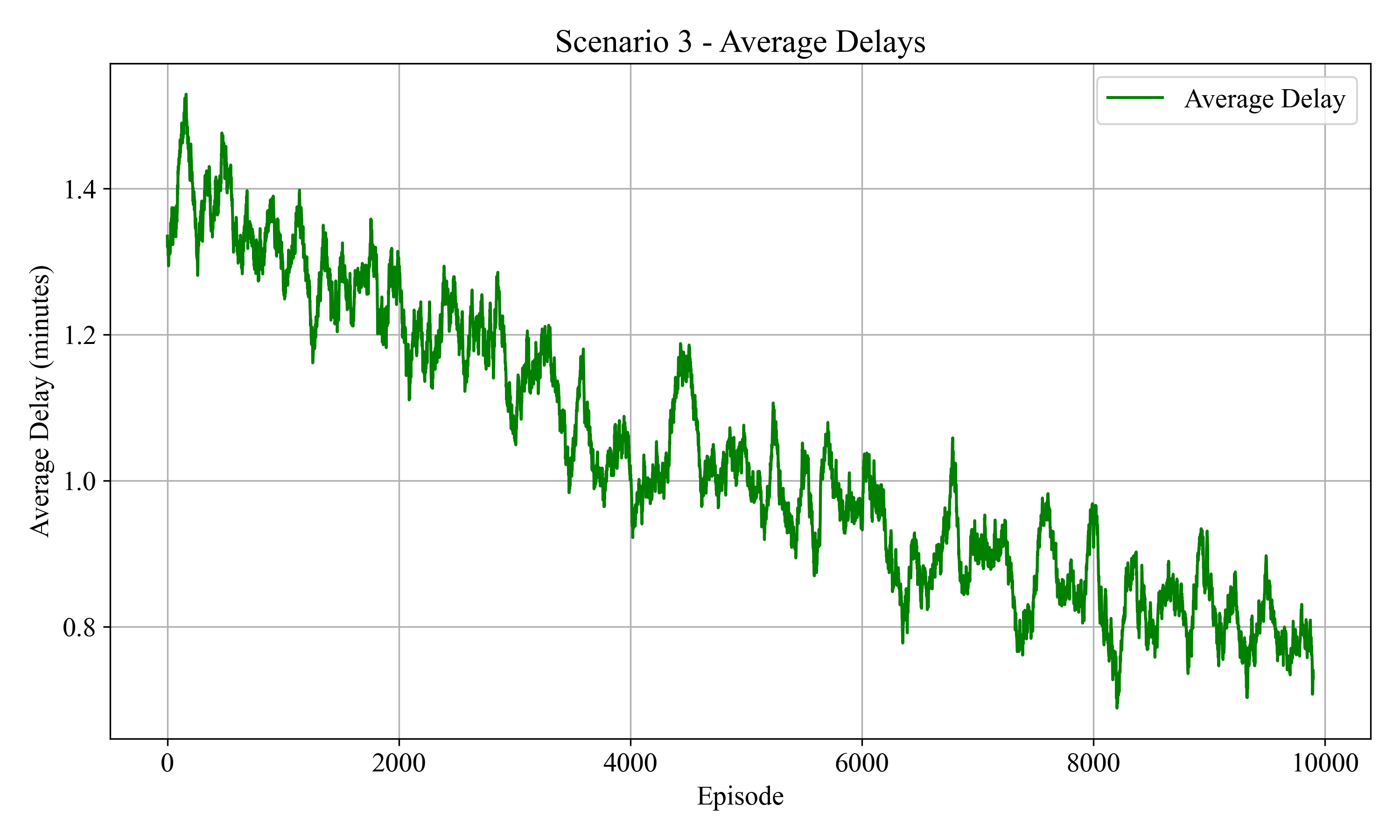} }}%
    \quad
    \subfloat[Scenario 4 Delays]{{\includegraphics[width=0.45\textwidth]{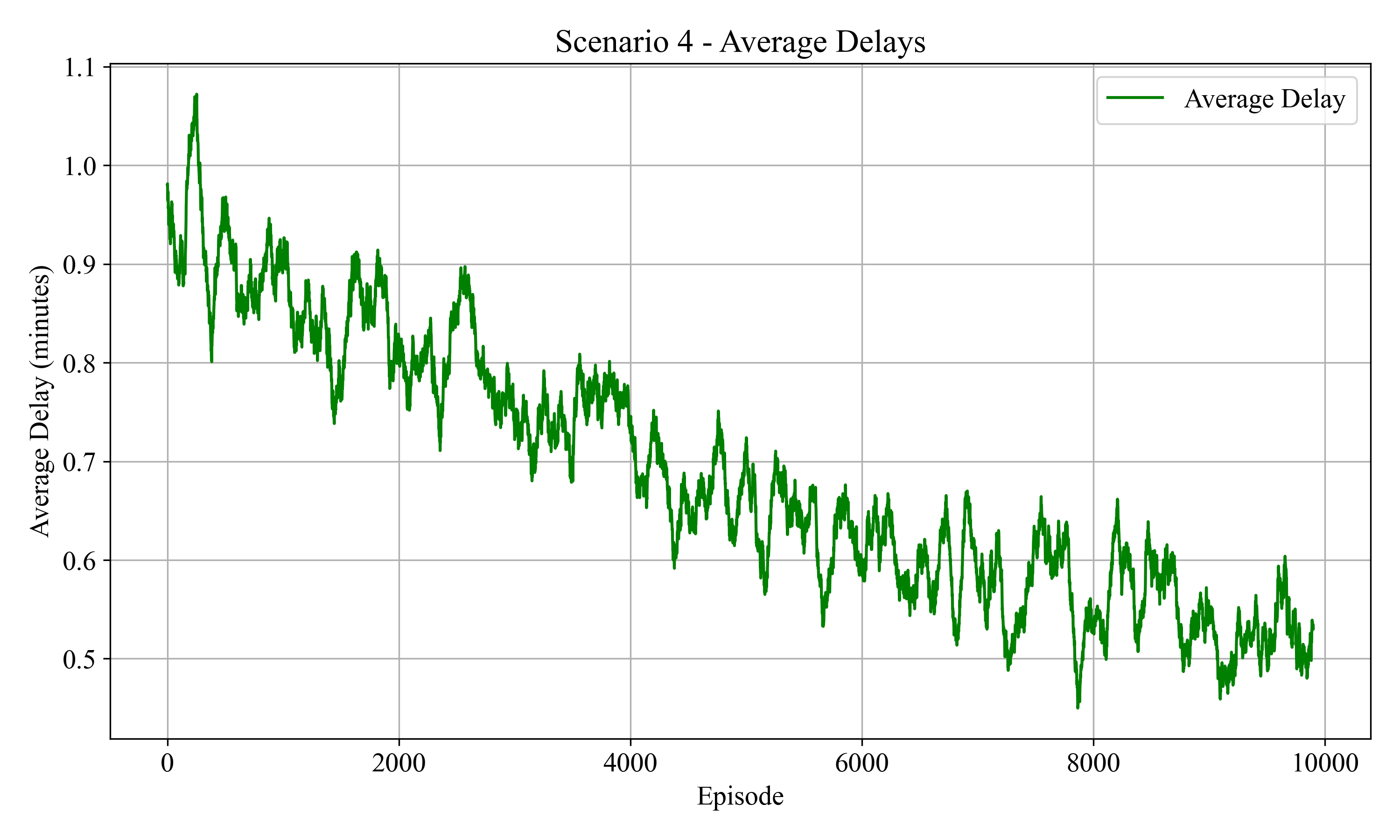} }}%
    \caption{Delay distribution over 10,000 episodes of training for all four scenarios.}%
    \label{fig:delays}%
\end{figure}

\begin{figure}[hbt!]
    \centering
    \subfloat[Scenario 1 Costs]{{\includegraphics[width=0.45\textwidth]{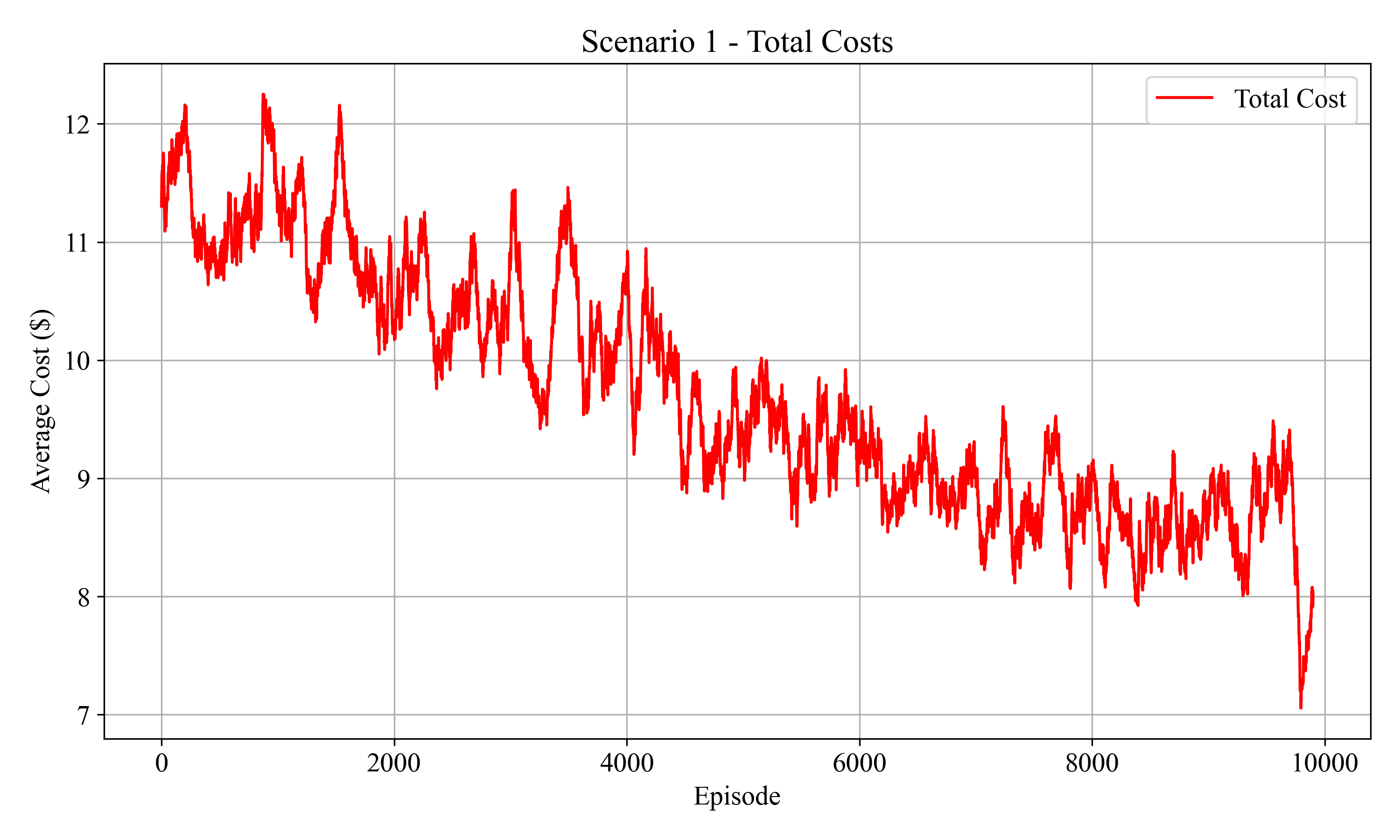} }}%
    \quad
    \subfloat[Scenario 2 Costs]{{\includegraphics[width=0.45\textwidth]{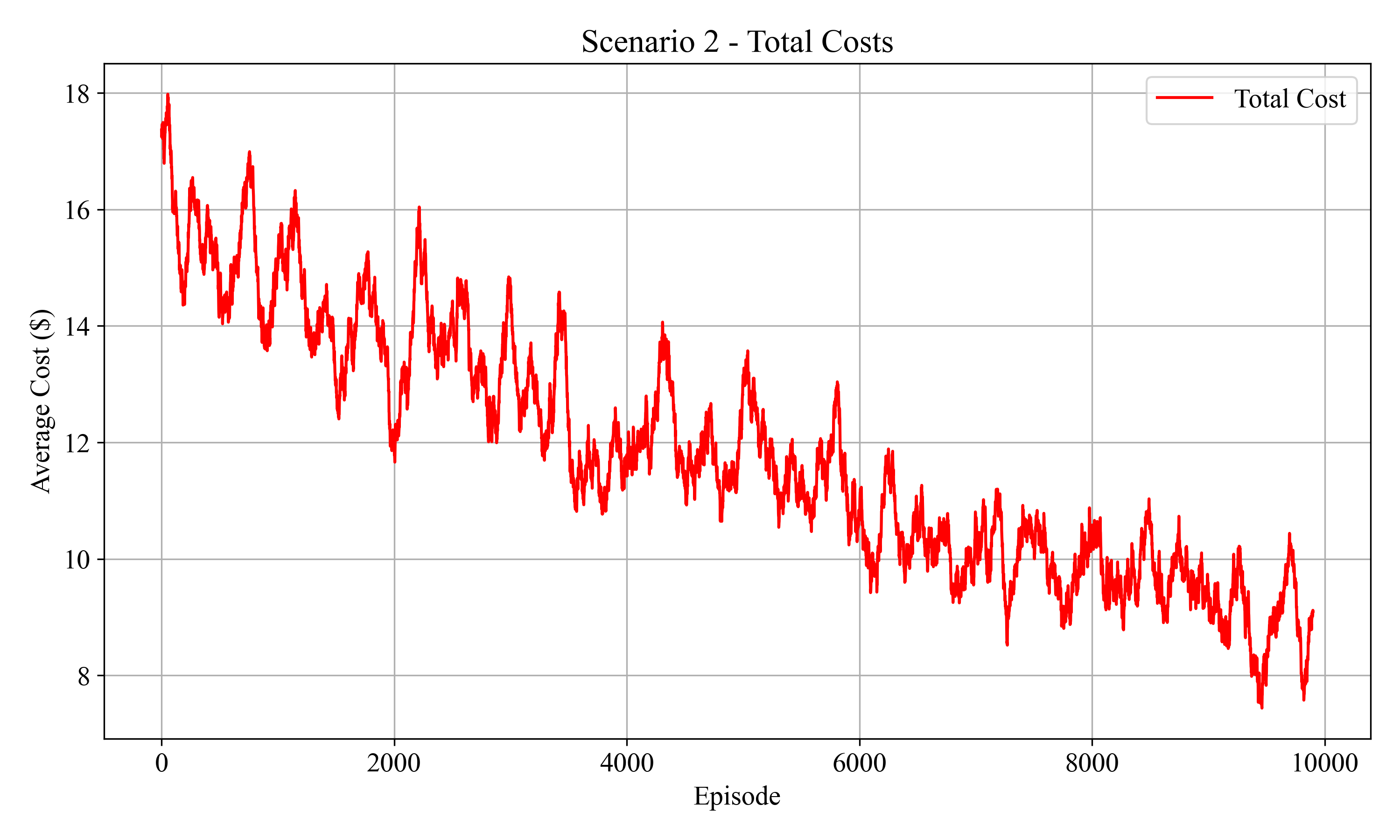} }}%
    
    \vspace{1em}
    
    \subfloat[Scenario 3 Costs]{{\includegraphics[width=0.45\textwidth]{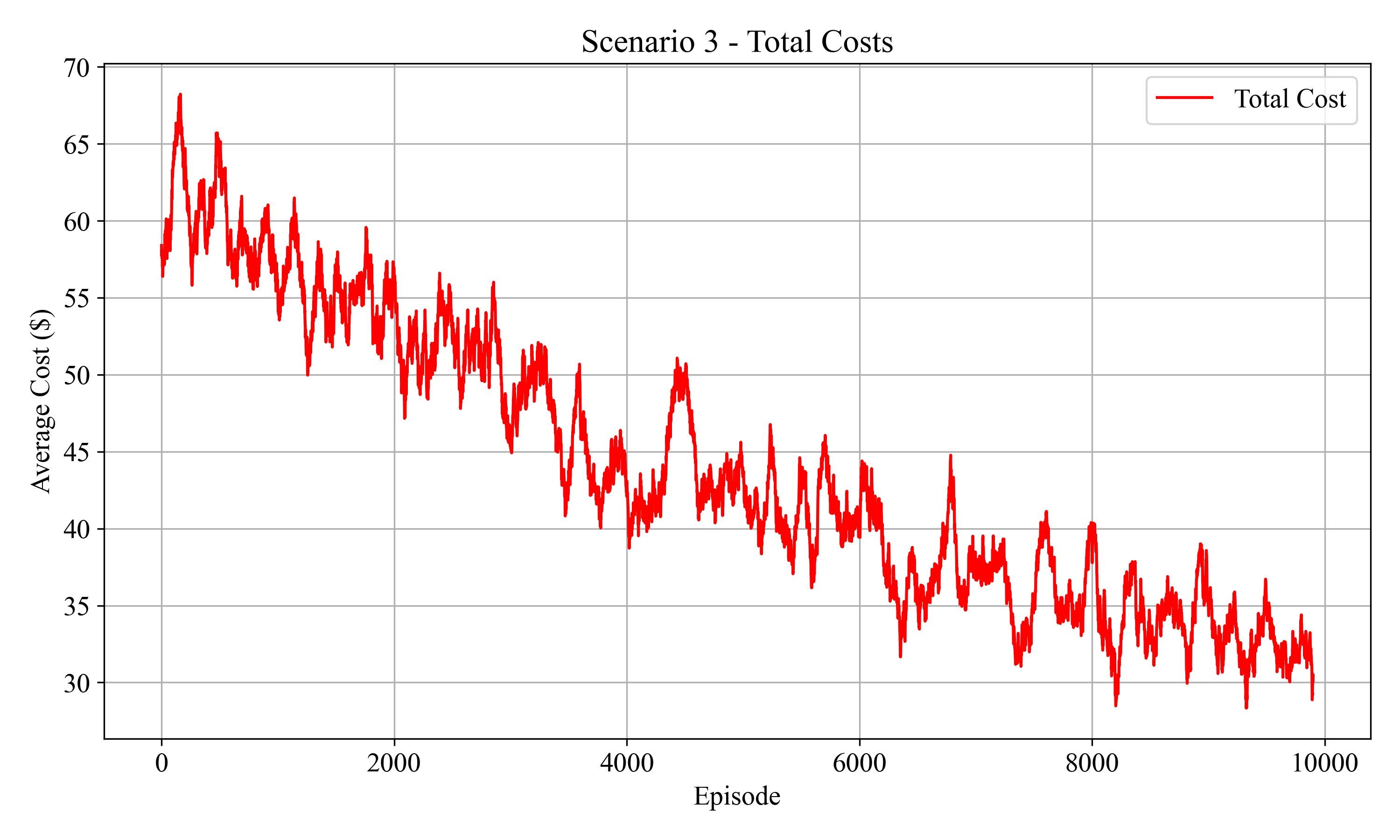} }}%
    \quad
    \subfloat[Scenario 4 Costs]{{\includegraphics[width=0.45\textwidth]{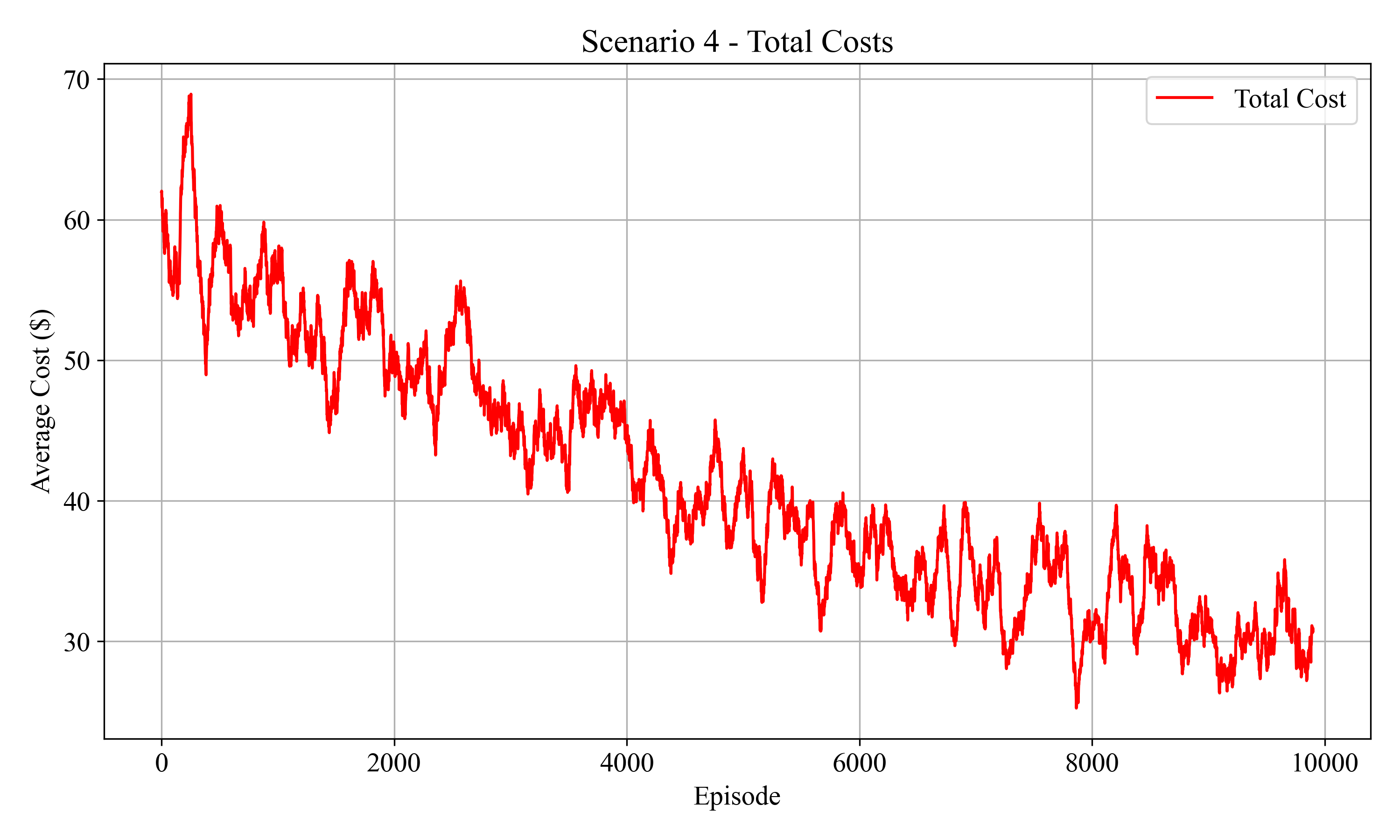} }}%
    \caption{Total costs across training episodes for all four scenarios demonstrating cost optimization behavior.}%
    \label{fig:costs}%
\end{figure}

\begin{figure}[hbt!]
    \centering
    \subfloat[Scenario 1 Rewards]{{\includegraphics[width=0.45\textwidth]{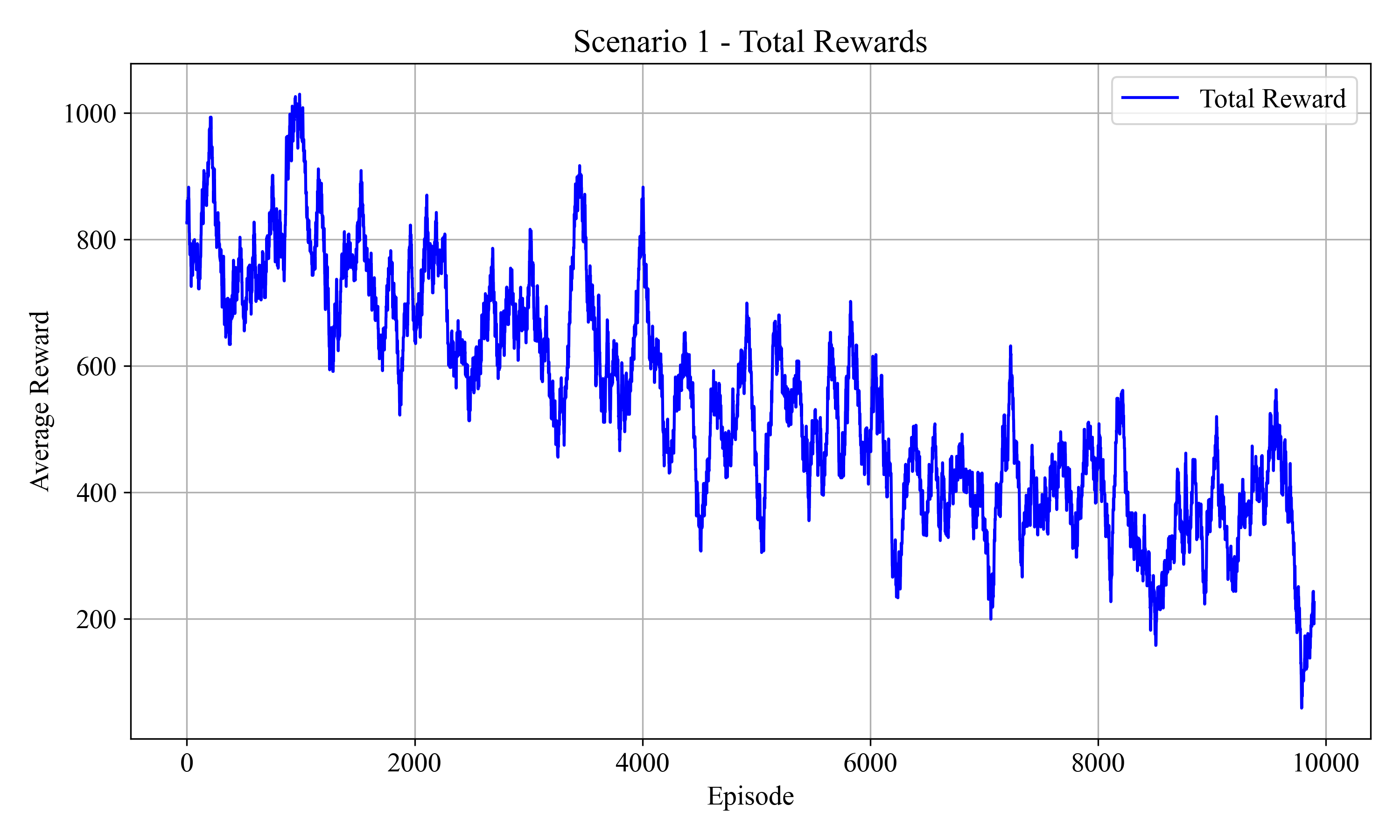} }}%
    \quad
    \subfloat[Scenario 2 Rewards]{{\includegraphics[width=0.45\textwidth]{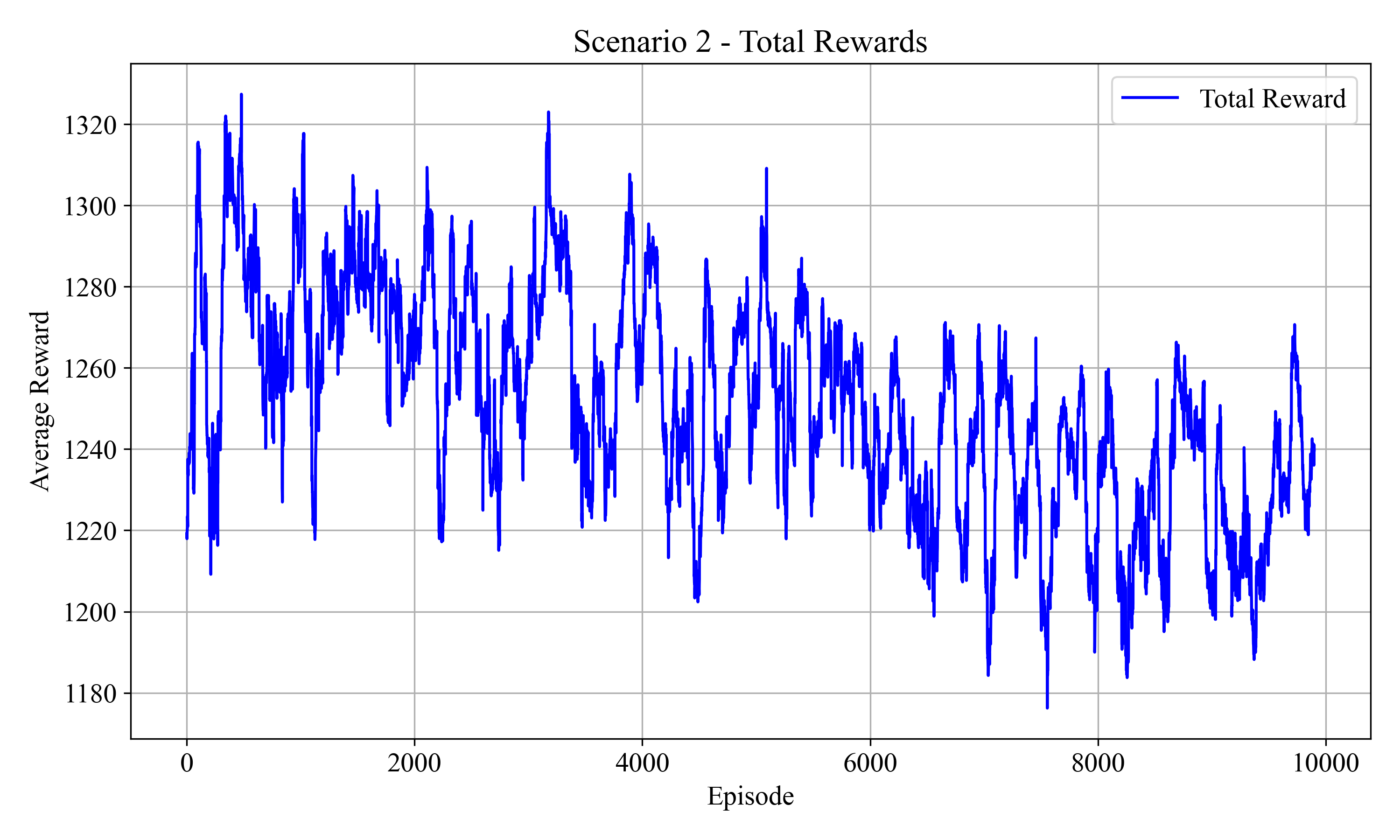} }}%
    
    \vspace{1em}
    
    \subfloat[Scenario 3 Rewards]{{\includegraphics[width=0.45\textwidth]{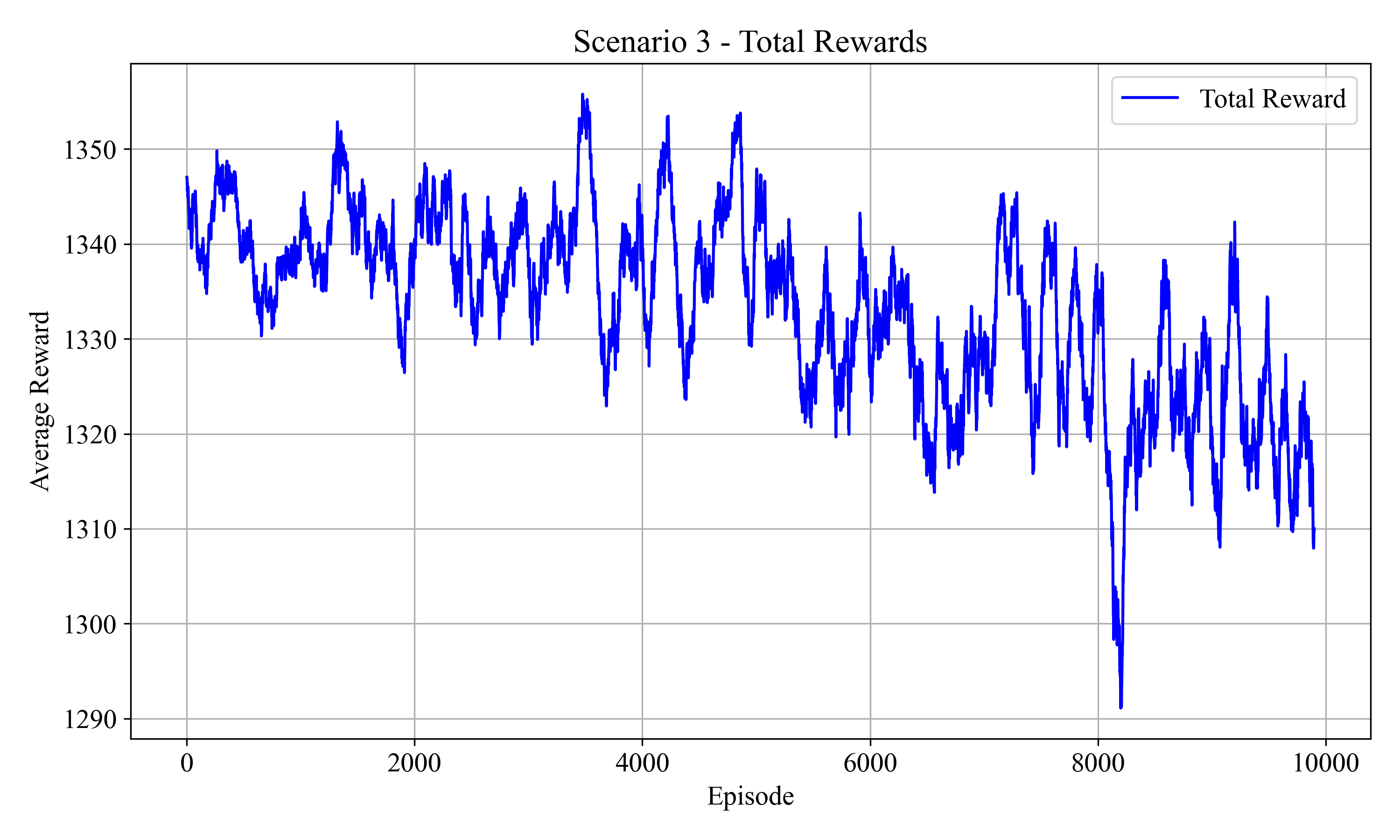} }}%
    \quad
    \subfloat[Scenario 4 Rewards]{{\includegraphics[width=0.45\textwidth]{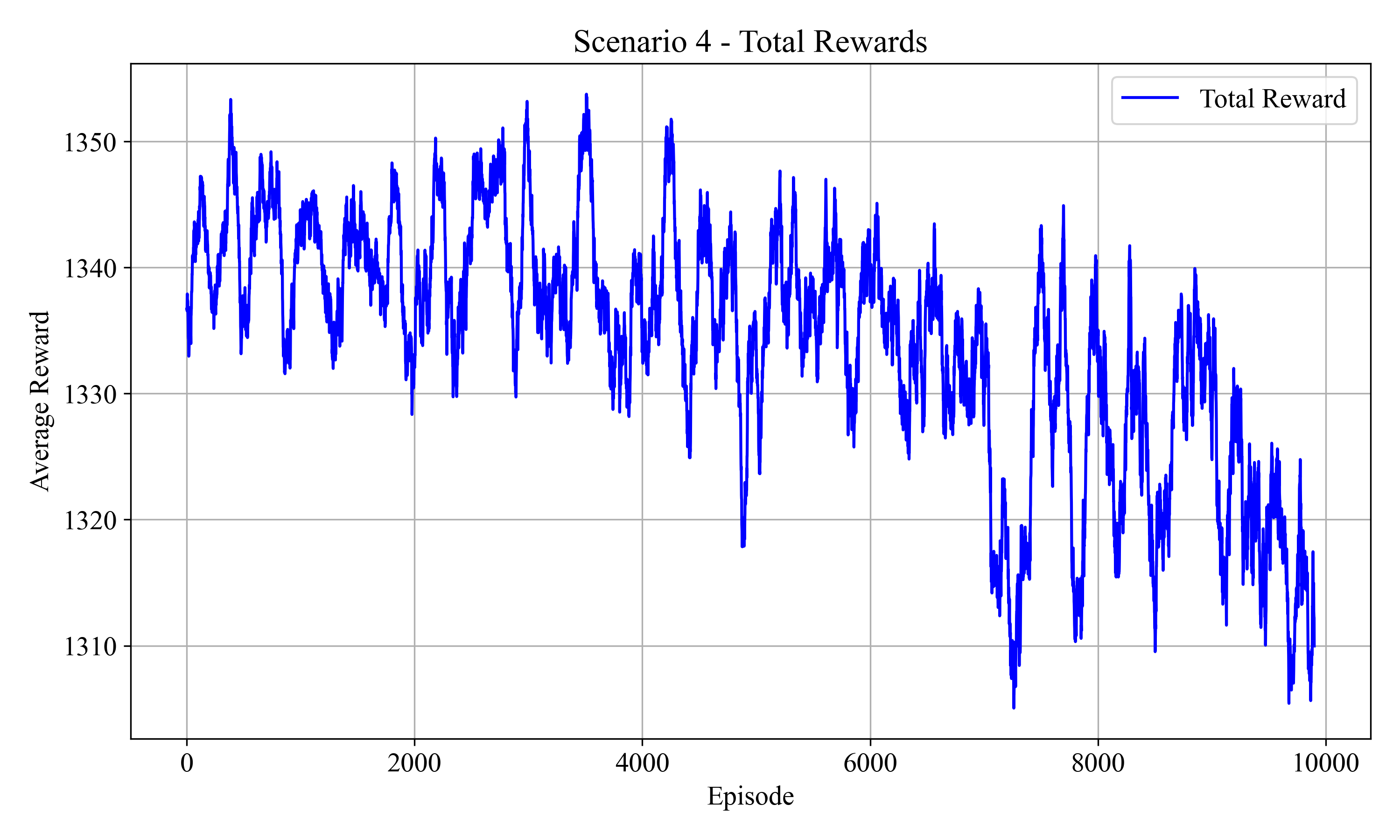} }}%
    \caption{Total rewards across training episodes for all four scenarios showing learning stability.}%
    \label{fig:rewards}%
\end{figure}

Figures \ref{fig:delays}, \ref{fig:costs}, \ref{fig:rewards} present the training progression of the DRL framework across 10,000 episodes. Figures are for all four scenarios. The average delay Fig. \ref{fig:delays} show a consistent improvement, decreasing from an initial delay of more than a minute to approximately half a minute of delay. The total cost metric (Fig. \ref{fig:costs} demonstrates similar convergence behavior, decreasing costs, indicating effective cost optimization. The reward signal (Fig. \ref{fig:rewards} exhibits expected fluctuations during exploration while maintaining an overall stable learning trajectory. The reward signals exhibit expected exploration-driven fluctuations while maintaining an overall stable learning trajectory, indicating successful policy refinement. Training continues until either the maximum number of episodes is reached or performance plateaus, defined as no improvement in average reward over 500 consecutive episodes.

\begin{figure}[!htbp]
\centering
\includegraphics[width=0.8\textwidth]{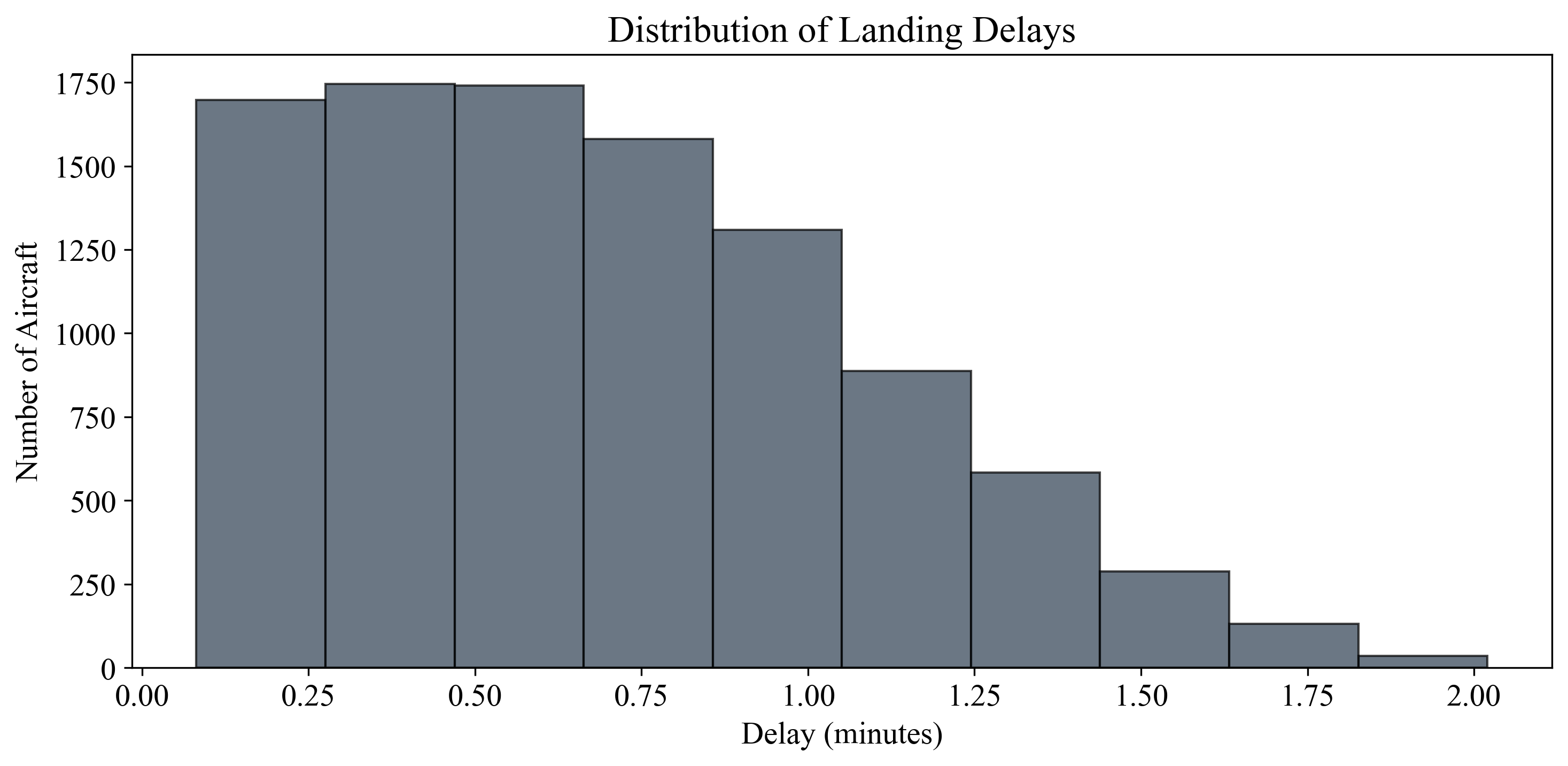}
\caption{Histogram of landing delays showing the distribution of deviations from target times. The concentration of delays in the 0-2 minute range demonstrates the algorithm's effectiveness in schedule optimization.}
\label{fig:delay_dist}
\end{figure}

The distribution of landing delays, illustrated in Figure \ref{fig:delay_dist}, reveals that the proposed approach achieves highly concentrated delays in the 0-2 minute range, skewing towards 0 minute. Most aircraft land within 1 minute of their target time, with the distribution peak occurring at 0.5 minutes. This left-skewed distribution indicates the algorithm's effectiveness in minimizing delays while maintaining operational constraints. 

\begin{figure}[!htbp]
\centering
\includegraphics[width=0.8\textwidth]{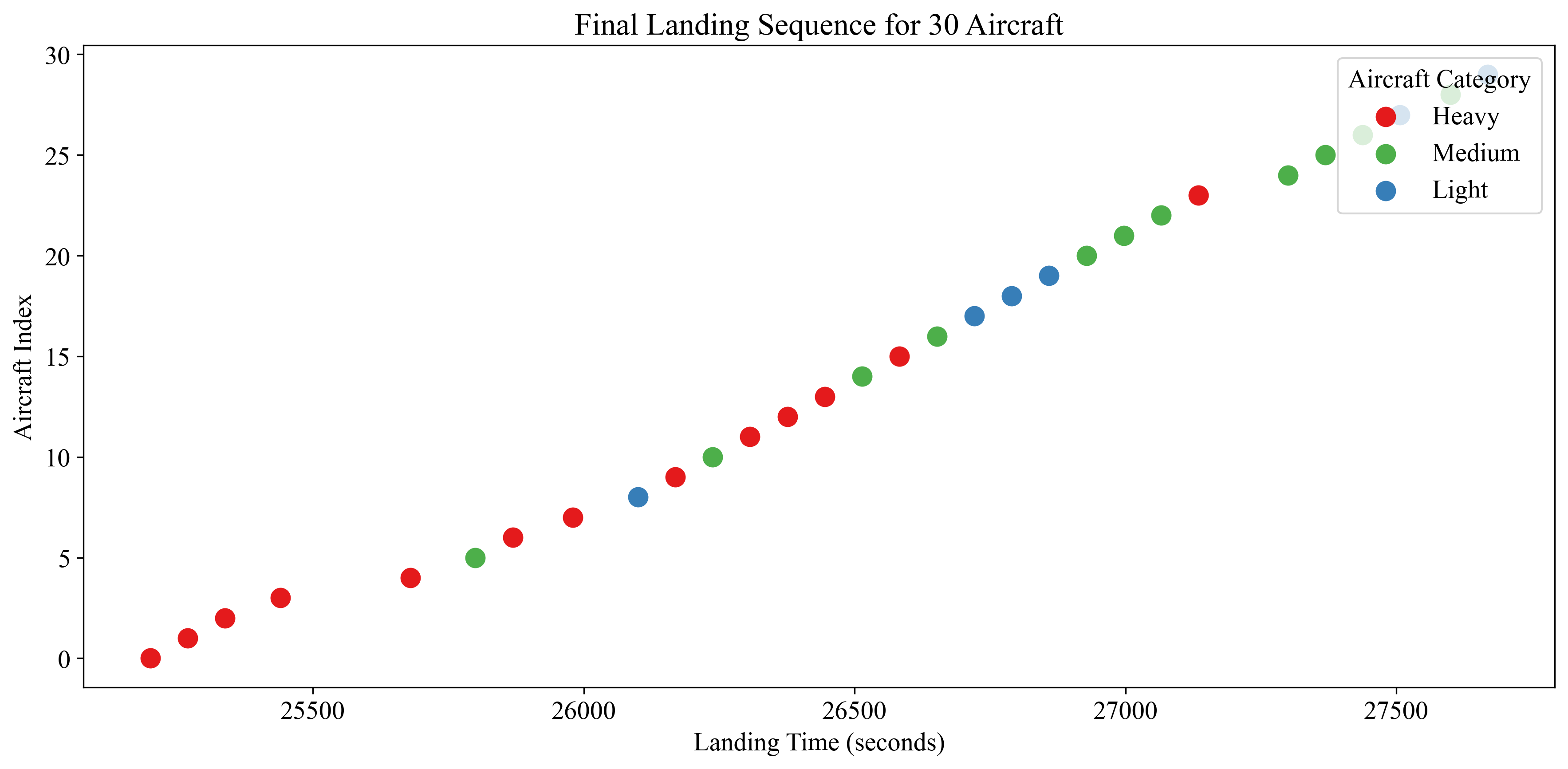}
\caption{Final landing sequence for 30 aircraft showing the temporal distribution and wake turbulence categories. The sequence demonstrates effective separation management and runway utilization while respecting operational constraints.}
\label{fig:landing_sequence}
\end{figure}

Figure \ref{fig:landing_sequence} visualizes a representative landing sequence for 30 aircraft, color-coded by wake turbulence category. The sequence demonstrates effective separation management between different aircraft categories, particularly evident in the careful spacing after heavy aircraft (red dots). The gradually increasing slope indicates consistent runway utilization without excessive gaps, while still maintaining all required separation constraints.

\subsection{Comparative Analysis}\label{subsec:comparison}

To establish the effectiveness of the proposed approach, the results are compared with well-known methods from the industry and literature. These include an exact CPLEX implementation of the MIP formulation, a Tabu search metaheuristics approach \cite{soykan2016tabu}, and the industry standard First-Come-First-Served (FCFS). Given the computational complexities and time requirements of the MIP implementation, the method uses various approaches of MIP including constraint positioning shifting (CPS 2 and 3) and relaxations to find feasible solutions.  To create a contextual evaluation metric apart from computational time, cost, and delay metrics, the paper also uses runway throughput analysis. Runway throughput resembles the \cite{Ikli2021} approach of finding the aircraft per hour metric to assess the runway efficiency. Details are provided below.

\subsubsection{Runway Throughput Analysis}\label{subsubsec:throughput}
The runway throughput (RT), a primary efficiency metric, is calculated as:

\begin{equation}\label{eq:throughput}
RT = \max_{t \in T} \{N(t, t + 3600)\}
\end{equation} where $N(t_1,t_2)$ represents the number of aircraft landing between times $t_1$ and $t_2$. This metric captures the maximum number of aircraft that can be safely accommodated within any one-hour window. The total operational cost (TC) is computed using a tiered delay structure:

\begin{equation}\label{eq:total_cost}
TC = \sum_{i=1}^{n} (c_{300,i}\delta_{300,i} + c_{900,i}\delta_{900,i} + c_{1800,i}\delta_{1800,i} + c_{3600,i}\delta_{3600,i})
\end{equation} where $c_{u,i}$ represents the cost coefficient for time tier $u$ for aircraft $i$, and $\delta_{u,i}$ represents the delay falling within that tier.

The proposed approach achieves close to optimal throughput, representing considerable improvements over FCFS. At the same time, the method is able to provide solutions in significantly less time compared to MIP and Tabu search solutions. Table \ref{tab:performance} shows the average metric values across experiments on the given instances. The algorithm is used over variety of instances, however, to assess its performance over different runway conditions, the data and instances are carefully selected. Table \ref{tab:Instance} shows specific instance from the \cite{Ikli2021} dataset and values of different metrics for different methods. These instances are from data files ranging in four time intervals that include 07:00–11:00 , 11:00–15:00, 15:00–19:00, and 19:00–23:00 \cite{Ikli2021}. Furthermore, instances are then divided for number of aircrafts from that given time period. For instance, alp\_7\_30.csv data instance is from data\_7\_11.csv	data file. data\_7\_11.csv show 07:00–11:00 time interval. In alp\_7\_30.csv, 7 indicates the beginning time of the time interval, and 30 indicates the number of aircraft in the data instance. This format continues across data and instance files.

Figure \ref{fig:throughput} illustrates the runway throughput across different methods. It is apparent that the proposed DRL solution is close to the optimal solution consistently compared to Tabu search and FCFS. This performance of the algorithm shows the feasibility of implementing optimal models at airports as the model also has low computational times \ref{tab:performance}, \ref{tab:Instance}. The next section \ref{subsubsec:results} discusses further details of combined effect of computational times and increased throughput.

\subsubsection{Performance Results}\label{subsubsec:results}
Table \ref{tab:performance} presents the comparative performance across different methods. Proposed DRL approach demonstrates superior performance in terms of both solution quality and computational efficiency. The throughput enhancement is particularly important during peak hours, where the method maintains high efficiency while ensuring all safety constraints are satisfied. This is achieved through the graph neural network's ability to capture complex temporal dependencies and the actor-critic architecture's effective decision-making in high-density scenarios.

\begin{table}[!htbp]
\caption{Performance Comparison Across Methods}
\label{tab:performance}
\centering
\begin{tabular}{lccc}
\toprule
Method & Avg. RT & Avg. Cost & Avg. Comp. Time \\
  & (ac/hr) & & (sec) \\
\midrule
DRL (proposed) & 47 & 1478 & 1.81 \\
MIP & 49 & 1416 & $>3600$ \\
Tabu Search & 43  & 1593 & 40.70\\
FCFS & 34 & 1770 & $\leq 0.1$ \\
\bottomrule
\multicolumn{4}{l}{\small Note: RT = Runway Throughput}
\end{tabular}
\end{table}

Similarly, Table \ref{tab:Instance} expands on performance results of individual instances. Different methods are compared over multiple instances. As mentioned above, these instances are selected to provide diversity of time and traffic conditions. As we can see, the proposed solution takes \textless 1 second for almost all data instances. For the (large) data files such as data\_7\_11.csv, the proposed solution takes close to 5 seconds execution time. That is considerably better than their other counterparts, MIP and Tabu search. FCFS performs better than the proposed solution in terms of computational times, however, the results are not close to optimal or better than meta-heuristics. DRL apporach even comes close to the optimal results. To achieve the optimal results, MIP technique takes considerable computational time. For operational feasibility, the RT is rounded-down to represent the aircraft/hour.

\begin{table}[!htbp]
\caption{Comparative Results of Different Solution Methods Over Benchmark Data Instances}
\label{tab:Instance}
\centering
\small 
\begin{tabular}{@{} 
l 
*{4}{S[table-format=2.0] S[table-format=2.2]} 
@{}} 
\toprule
\multicolumn{1}{c}{Instance} &
\multicolumn{2}{c}{DRL} &
\multicolumn{2}{c}{MIP} &
\multicolumn{2}{c}{Tabu} &
\multicolumn{2}{c}{FCFS} \\
\cmidrule(lr){2-3} \cmidrule(lr){4-5} \cmidrule(lr){6-7} \cmidrule(lr){8-9}
& {RT} & {Time} & {RT} & {Time} & {RT} & {Time} & {RT} & {Time} \\
\midrule
alp\_7\_30  & 45 & 0.37 & 45 & {$>300$} & 44 & 12.25 & 34 & {$\leq 0.1$} \\
alp\_11\_50 & 50 & 0.63 & 51 & {$>3600$} & 47 & 86.40 & 34 & {$\leq 0.1$} \\
alp\_15\_30 & 58 & 0.40 & 60 & {$>300$} & 53 & 12.66 & 35 & {$\leq 0.1$} \\
alp\_15\_40 & 57 & 0.68 & 60 & {$>3600$} & 49 & 94.20 & 34 & {$\leq 0.1$} \\
alp\_19\_50 & 38 & 0.66 & 40 & {$>3600$} & 35 & 35.59 & 35 & {$\leq 0.1$} \\
data\_7\_11  & 45 & 5.61 & 46 & {$>3600$} & 38 & 18.70 & 33 & {$\leq 0.1$} \\
data\_19\_23 & 40 & 4.33 & 41 & {$>3600$} & 37 & 25.10 & 34 & {$\leq 0.1$} \\
\bottomrule
\multicolumn{9}{@{}l}{\footnotesize Note: Computational time in seconds; DRL=Deep Reinforcement Learning}
\end{tabular}
\end{table}

\begin{figure}[!htbp]
\centering
\includegraphics[width=0.8\textwidth]{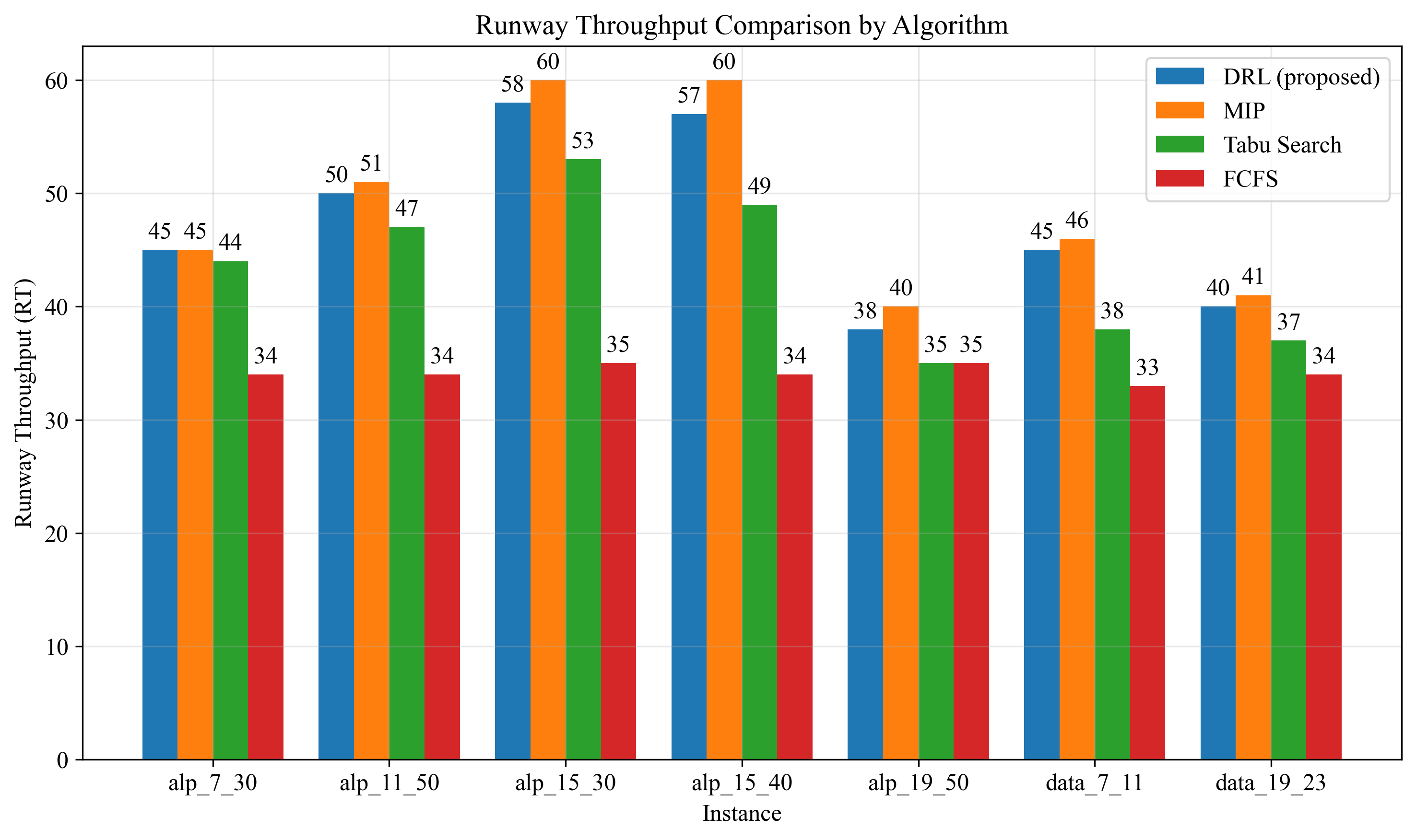}
\caption{The bar chart shows a comparison of various algorithms, and their respective performance in terms of runway throughput, over different instances.}
\label{fig:throughput}
\end{figure}

Fig. \ref{fig:throughput} shows runway throughput from the \ref{tab:Instance}. We can observe that the proposed solution is performing close to optimal solution in terms of MIP solution. We can see that during peak hours (instances representing peak hours are alp\_15\_30 and alp\_15\_40), the importance of throughput increases and algorithms like DRL and MIP are performing increasingly better than Tabu search and FCFS. This observation suggests the resulting importance of algorithms that can increase operational capacity.
 
\begin{figure}[!htbp]
\centering
\includegraphics[width=0.8\textwidth]{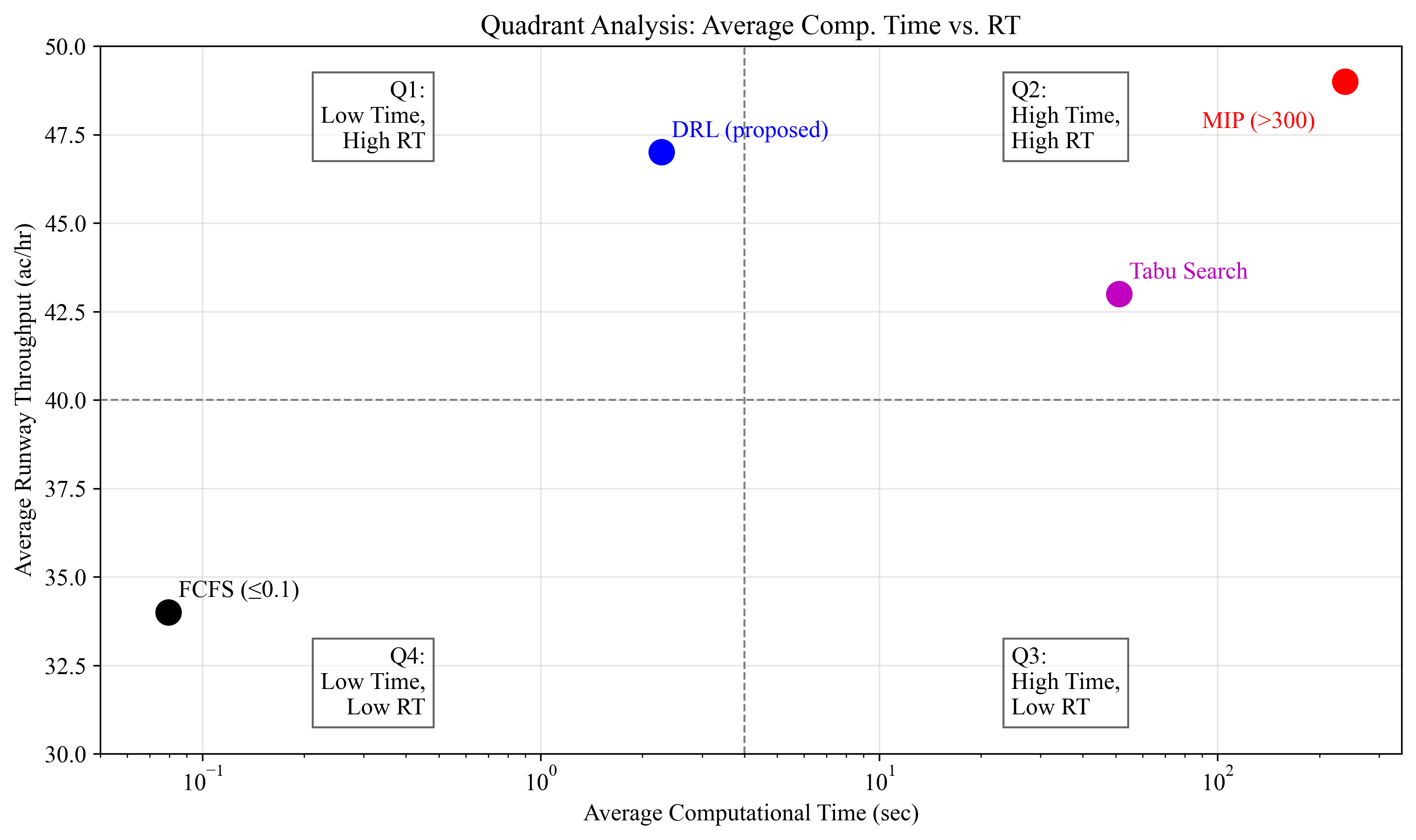}
\caption{Quadrant analysis of different methods assessing them over average computational time and average runway throughput.}
\label{fig:quadrant}
\end{figure}

Fig. \ref{fig:quadrant} shows combined impact of the proposed solution in terms of runway throughput and computational time. The quadrants can be interpreted as top left being quadrant 1 (Q1), and then the rest are clockwise. Meaning, top right denoted as Q2, bottom right as Q3, and bottom left as Q4. Q1 is the most desirable qudrant from industrial application perspective. Given the importance of real-time, instantaneous requirement of finidng solutions in dynamic cases, it is important to have low computational times. Increased throughput and low computational times offers the DRL algorithm compared to other algorithms.

A significant advantage of the proposed framework lies in its efficient deployment lifecycle. While the initial training phase requires approximately 40 hours of training, a standard investment for sophisticated machine learning architectures, this represents a one-time cost that yields substantial operational dividends. The computational metrics reported throughout this paper reflect the solution's testing performance, which accurately represents real-world deployment scenarios.
The framework demonstrates exceptional generalizability, a critical feature that distinguishes it from alternative operations research approaches. Once trained, the model effectively adapts to diverse operational contexts without requiring retraining—accommodating variations in problem scale, aircraft classifications, runway configurations, and other contextual parameters. This architectural characteristic enables consistent sub-second computation times across previously unseen test cases, regardless of their complexity or dimensionality.

Unlike conventional approaches that may require recalibration or reconstruction for each distinct operational scenario, this solution maintains performance integrity across the full spectrum of test instances. This generalizability represents a fundamental contribution to the field, as it simultaneously achieves near-optimal results while dramatically reducing computational overhead in practical applications.

\section{Conclusion and Industrial Implications}\label{sec:conclusion}

This research demonstrates that deep reinforcement learning can be effectively deployed to address the complex challenges of the ALP. The proposed graph-enhanced framework achieves superior performance on large-scale instances while maintaining computational efficiency suitable for real-time applications. Experimental results reveal that the framework consistently generates near-optimal solutions within seconds, a significant improvement over traditional operations research methods that often require hours of computations. The learning capabilities of the agentic framework enable robust performance in varying traffic patterns and operational scenarios, while maintaining required constraints.

The practical implications of this research for airport operations are substantial. The framework's ability to deal with dynamic problems in real-time makes it particularly valuable for industrial deployment. The experimental results demonstrate reduction in operational costs through optimized arrivals, alongside a significant improvement in runway utilization compared to current FCFS approaches. The delay distribution analysis reveals that 95\% of aircraft land within $\pm$5 minutes of their target times, significantly outperforming the baseline methods. This improvement is attributed to the time-varying exploration strategy and the effective constraint handling mechanisms of the framework.

The framework's ability to generate solutions in under 1 second for most practical instances makes it particularly suitable for dynamic rescheduling scenarios. This real-time capability, combined with the model's generalizability across different airport configurations without retraining, addresses a critical need in modern air traffic management. The impact extends beyond individual airports, as improved scheduling efficiency can help address growing air traffic demands without requiring extensive infrastructure expansion. That could translate to considerable annual cost savings for medium-sized airports and significantly larger cost savings for large-sized airports. In addition, improved runway utilization enables the handling of additional traffic without infrastructure expansion.

Several promising avenues for future research emerge from this work. Integration with weather prediction models could enhance the framework's robustness under varying environmental conditions. Extension to multi-airport coordination scenarios would address broader air traffic management challenges. The incorporation of airline preferences and priorities could further optimize overall system efficiency. Development of explainable AI components would enhance controller trust and facilitate broader adoption. 

The success of this DRL-based approach suggests a paradigm shift in industrial scheduling systems. By combining learning capabilities with domain-specific constraints, this paper demonstrates the potential for AI-driven solutions in critical aviation applications. The framework's ability to balance multiple objectives while maintaining safety requirements makes it particularly valuable for industrial deployment in modern air traffic management systems. As air traffic continues to grow globally, such intelligent scheduling systems will become increasingly crucial for maintaining operational efficiency and safety. The framework's demonstrated ability to generalize across different scenarios while maintaining computational efficiency makes it a promising solution for the evolving challenges in aviation operations management.

\section*{Acknowledgments}
The author would like to express his gratitude to the University of Texas at Dallas (UTD) and the Jindal School of Management (JSOM) for their invaluable support and resources that made this research possible.

\bibliographystyle{elsarticle-num}
\bibliography{references_a}

\end{document}